\documentclass[acmtog,nonacm,screen]{acmart}

\let\titleold\title
\renewcommand{\title}[1]{\titleold{#1}\newcommand{\thetitle}{#1}}
\def\maketitlesupplementary
   {
   \newpage
       \twocolumn[
        \centering
        \Large
        \textbf{\thetitle}\\
        \vspace{0.5em}Supplementary Material \\
        \vspace{1.0em}
       ] 
   }
   
\usepackage{booktabs} 

\usepackage{array} 
\usepackage{bm}
\usepackage{amsfonts}
\usepackage{threeparttable}
\usepackage{graphicx}
\usepackage{color}
\usepackage{colortbl}
\usepackage{arydshln}
\usepackage{enumitem}
\usepackage{multirow}
\usepackage{cleveref}
\usepackage{enumitem}
\makeatletter
\DeclareRobustCommand\onedot{\futurelet\@let@token\@onedot}
\def\@onedot{\ifx\@let@token.\else.\null\fi\xspace}

\def\eg{\emph{e.g}\onedot}

\def\etal{\emph{et al}\onedot}
\makeatother

\usepackage[ruled]{algorithm2e} 

\SetAlFnt{\small}
\SetAlCapFnt{\small}
\SetAlCapNameFnt{\small}
\SetAlCapHSkip{0pt}

\acmJournal{TOG}

\begin{document}
\title{Deformable Beta Splatting}

\author{Rong Liu}
\email{roliu@ict.usc.edu}
\authornote{Co-first authors, equal technical contribution.}
\affiliation{%
  \institution{University of Southern California, Institute for Creative Technologies}
  \city{Los Angeles}
  \state{California}
  \country{USA}
}
\author{Dylan Sun}
\email{dylansun@usc.edu}
\authornotemark[1]
\affiliation{%
  \institution{University of Southern California}
\city{Los Angeles}
  \state{California}
  \country{USA}
}
\author{Meida Chen}
\email{mechen@ict.usc.edu}
\affiliation{%
  \institution{University of Southern California, Institute for Creative Technologies}
  \city{Los Angeles}
  \state{California}
  \country{USA}
}
\author{Yue Wang}
\email{yue.w@usc.edu}
\authornote{Co-advisors, equal leading contribution.}
\affiliation{%
  \institution{University of Southern California}
  \city{Los Angeles}
  \state{California}
  \country{USA}
}
\author{Andrew Feng}
\email{feng@ict.usc.edu}
\authornotemark[2]
\affiliation{%
  \institution{University of Southern California, Institute for Creative Technologies}
  \city{Los Angeles}
  \state{California}
  \country{USA}
}

\begin{abstract}
3D Gaussian Splatting (3DGS) has advanced radiance field reconstruction by enabling real-time rendering. However, its reliance on Gaussian kernels for geometry and low-order Spherical Harmonics (SH) for color encoding limits its ability to capture complex geometries and diverse colors.
We introduce Deformable Beta Splatting (DBS), a deformable and compact approach that enhances both geometry and color representation. DBS replaces Gaussian kernels with deformable Beta Kernels, which offer bounded support and adaptive frequency control to capture fine geometric details with higher fidelity while achieving better memory efficiency. In addition, we extended the Beta Kernel to color encoding, which facilitates improved representation of diffuse and specular components, yielding superior results compared to SH-based methods. Furthermore, Unlike prior densification techniques that depend on Gaussian properties, we mathematically prove that adjusting regularized opacity alone ensures distribution-preserved Markov chain Monte Carlo (MCMC), independent of the splatting kernel type. Experimental results demonstrate that DBS achieves state-of-the-art visual quality while utilizing only 45\% of the parameters and rendering 1.5x faster than 3DGS-MCMC, highlighting the superior performance of DBS for real-time radiance field rendering. 
Interactive demonstrations and source code are available on our project website:
\href{https://rongliu-leo.github.io/beta-splatting/}{https://rongliu-leo.github.io/beta-splatting/}.
\end{abstract}

%
%

\ccsdesc[500]{Computing methodologies~Rasterization}
\ccsdesc[500]{Computing methodologies~Shape representations}
\ccsdesc[500]{Computing methodologies~Reflectance modeling}
\ccsdesc[500]{Computing methodologies~Machine learning approaches}

%
%

\keywords{Novel View Synthesis, Radiance Fields, 3D Ellipsoidal Betas, Spherical Betas}

\begin{teaserfigure}
  \includegraphics[width=0.95\textwidth]{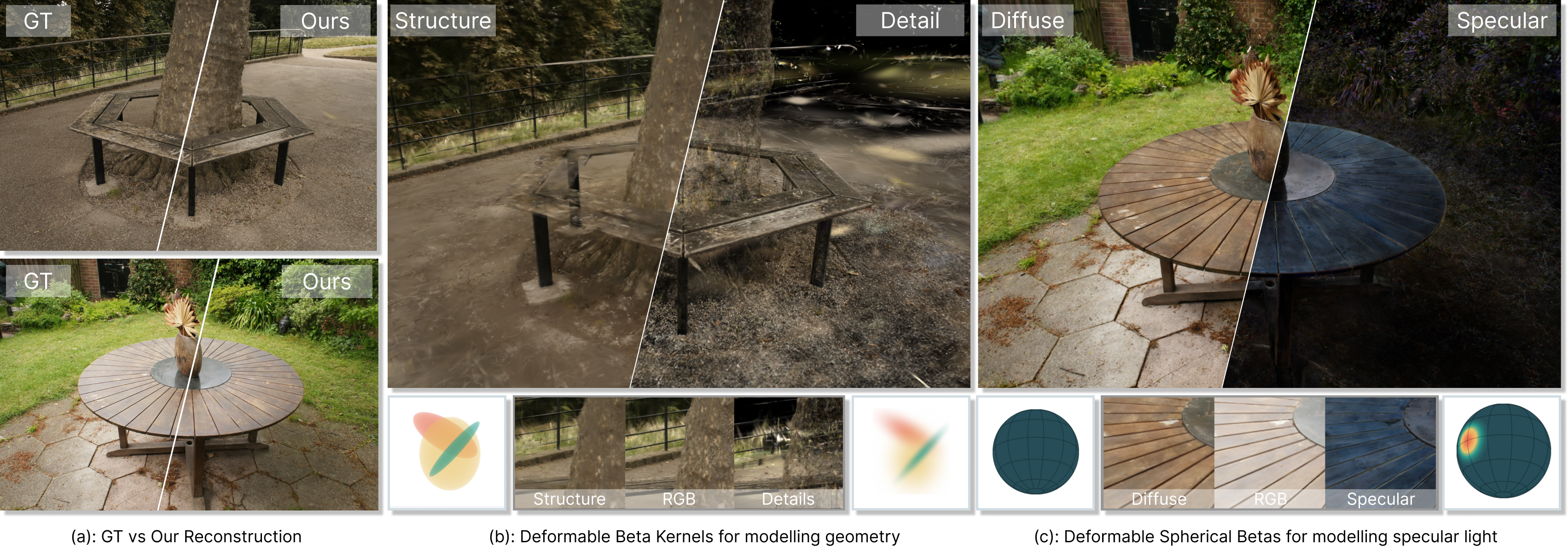}
  \caption{Superior scene reconstruction by Beta Splatting. Our deformable Beta Kernels achieve high-fidelity geometry and appearance representation (a), while naturally enabling the decoupling of surface geometry from texture details (b) and the separation of specular and diffuse lighting components (c). }
  \label{fig:teaser}
\end{teaserfigure}

\maketitle

\section{Introduction}
\label{sec:intro}
Neural Radiance Fields (NeRF)~\cite{mildenhall2021nerf,barron2021mipnerf,M_ller_2022,barron2023zipnerfantialiasedgridbasedneural} have demonstrated exceptional capabilities in 3D scene representation and novel view synthesis but are hindered by high computational costs and limited applicability in real-time scenarios due to their reliance on neural network-based encoding and dense volumetric ray-marching.
More recently, 3D Gaussian Splatting (3DGS)~\cite{3dgs} introduced an explicit representation paradigm that shifts from volumetric ray-marching to point-based splatting. By representing scenes as a set of 3D Gaussian primitives, 3DGS significantly reduces computational overhead, achieving real-time radiance field rendering while maintaining visual quality comparable to earlier NeRF methods. This explicit representation eliminates the need for heavy neural networks and enables direct manipulation of scene geometry and real-time rendering, making 3DGS ideal for applications such as immersive VR exploration~\cite{jiang2024vrgsphysicaldynamicsawareinteractive, franke2024vrsplattingfoveatedradiancefield, yu2024densoftdensespaceorientedlight}, dynamic scene representation~\cite{yang2023deformable3dgaussianshighfidelity, luiten2023dynamic3dgaussianstracking, wu20244dgaussiansplattingrealtime, yang2024realtimephotorealisticdynamicscene, guo2024motionaware3dgaussiansplatting}, 3D content creation~\cite{tang2024dreamgaussiangenerativegaussiansplatting, yi2024gaussiandreamerfastgenerationtext,chung2023luciddreamerdomainfreegeneration3d, yi2024gaussiandreamerprotextmanipulable3d}, mesh extraction~\cite{guédon2023sugarsurfacealignedgaussiansplatting, 2dgs, liu2024atomgsatomizinggaussiansplatting,dai2024highqualitysurfacereconstructionusing}, semantic field distillation~\cite{shi2023languageembedded3dgaussians, zhou2024feature3dgssupercharging3d, qin2024langsplat3dlanguagegaussian, qiu2024featuresplattinglanguagedrivenphysicsbased, ye2024gaussiangroupingsegmentedit}, and cinematic production~\cite{wang2024cinematic}.  
However, while 3DGS excels in efficiency, its rendering quality still falls short of the latest NeRF advancements~\cite{barron2023zipnerfantialiasedgridbasedneural}, particularly in capturing fine details and modeling view-dependent colors. To address this gap, a range of extensions has emerged, focusing on improving its visual quality through anti-alising~\cite{yu2023mipsplattingaliasfree3dgaussian,yan2024multiscale3dgaussiansplatting,liang2024analyticsplattingantialiased3dgaussian,song2024sagsscaleadaptivegaussiansplatting}, deblurring~\cite{lee2024deblurring3dgaussiansplatting, peng2024bagsbluragnosticgaussian, zhao2024badgaussiansbundleadjusteddeblur, seiskari2024gaussiansplattingmoveblur, oh2024deblurgsgaussiansplattingcamera}, appearance modeling~\cite{xu2024texturegsdisentanglinggeometrytexture, svitov2024billboardsplattingbbsplatlearnable, weiss2024gaussianbillboardsexpressive2d, yang2024specgaussiananisotropicviewdependentappearance}, and revisting densification~\cite{lu2023scaffoldgsstructured3dgaussians,niemeyer2024radsplatradiancefieldinformedgaussian,zhang2024pixelgsdensitycontrolpixelaware,liu2024atomgsatomizinggaussiansplatting,bulò2024revisingdensificationgaussiansplatting,kheradmand20243dgaussiansplattingmarkov}. Despite these efforts, 3DGS-based methods have yet to surpass the state-of-the-art NeRF methods~\cite{kulhanek2024nerfbaselines}.

The limitations of 3DGS primarily arise from the constraints of its foundational components. The Gaussian kernel, while enabling real-time rendering through its explicit representation, inherently produces smooth and blurred results. This smoothing effect hinders its ability to accurately depict flat surface geometries and sharp edges, resulting in a loss of fine detail and diminished realism in rendered scenes. Additionally, the Gaussian kernel’s long-tailed and unbounded nature necessitates a hard-coded cut-off for efficient rendering. This workaround introduces cut-off artifacts and further compromises the accurate representation of fine geometric details.
Furthermore, 3DGS employs low--order Spherical Harmonics (SH) to encode view-dependent color information. Although SH can model lighting variations, its parameter count increases polynomially with the SH level, limiting practical use in real-time applications. As a result, 3DGS relies on lower SH levels, which inherently produce smooth and overly simplified color transitions. This prevents the accurate modeling of specular highlights and fails to decouple diffuse and specular components, resulting in suboptimal representations of complex lighting interactions. Notably, both Gaussian kernel and SH are fixed-function mappings, restricting the adaptability required to handle varying scene complexities and lighting conditions, further compromising visual fidelity. 

To address these shortcomings, we propose Deformable Beta Splatting (DBS), a novel framework designed to enhance both geometric and color representation in radiance field rendering. At the core of DBS is the Beta Kernel, inspired by the Beta distribution, which provides bounded support and flexible shape control. This kernel adapts naturally to diverse scene complexities, enabling accurate representation of flat surfaces, sharp edges, and smooth regions with high fidelity.
For color representation, we introduce the Spherical Beta (SB) function for view-dependent color encoding. By decoupling diffuse and specular components, SB effectively models sharp specular reflections and captures high-frequency lighting effects.
Compared to spherical harmonics (SH) of degree 3, SB achieves superior performance using only 31\% of the parameters, significantly improving model efficiency.
Finally, we revisit and adapt Markov Chain Monte Carlo (MCMC) principles from 3DGS-MCMC~\cite{kheradmand20243dgaussiansplattingmarkov} into the Deformable Beta Splatting framework to address unique challenges posed by deformable kernels. 
Specifically, the adaptive density control in 3DGS and the probability-preserved densification strategies in 3DGS-MCMC were designed for Gaussian kernels, making their direct application to deformable Beta kernels nontrivial. To overcome this, we mathematically prove that by regularizing opacity, adjusting opacity alone is valid for distribution-preserved densification, regardless of the number of densifications or the choice of splatting kernel. This results in a kernel-agnostic MCMC strategy, eliminating the need for complex heuristics and scale recomputation, thereby streamlining both the optimization and densification processes. Additionally, we redefine the noise function used in MCMC with the Beta kernel to maintain mathematical consistency and compatibility with opacity constraints.


The comprehensive experimental results demonstrate that our approach consistently outperforms both 3DGS-based and NeRF-based methods in visual quality across all benchmarks. Furthermore, our method utilizes only 45\% of the parameters required by 3DGS and renders 1.5x faster than 3DGS-MCMC, demonstrating DBS's superior performance in real-time radiance field rendering. 

To summarize, the main contributions of our paper are as follows: 

\begin{enumerate}
    \item Beta Kernel for Geometry Representation: A bounded, flexible kernel that improves geometric fidelity by accurately capturing flat surfaces, sharp edges, and smooth regions. 
    \item Spherical Beta for Color Encoding: An efficient view-dependent color model that decouples diffuse and specular components, enabling sharp specular highlights with fewer parameters. 
    \item Kernel-Agnostic MCMC for optimization: A distribution-preserved densification leveraging regularized opacity and a redefined Beta noise function, together enhancing optimization stability and efficiency independent of kernel shape.
\end{enumerate}

\section{Related Work} \label{sec:relatedwork}



This section reviews key approaches in novel view synthesis, including implicit NeRFs and explicit 3DGS, highlighting how our work builds on these advancements and contributes to high-fidelity, real-time 3D rendering.

\subsection{Neural Radiance Fields}

The introduction of Neural Radiance Fields (NeRFs)~\cite{mildenhall2021nerf} revolutionized 3D scene representation and novel view synthesis. By encoding a continuous volumetric scene within a multilayer perceptron (MLP), NeRFs jointly learn scene density and view-dependent color, enabling photorealistic novel view synthesis through differentiable volumetric rendering.

Since their inception, NeRFs have spurred extensive research to address challenges in convergence speed, computational efficiency, and scalability. Early advancements, such as FastNeRF~\cite{garbin2021fastnerf} and PlenOctrees~\cite{yu2021plenoctrees}, improved efficiency through caching strategies and spatial acceleration structures. Methods like NeRF++\cite{zhang2020nerf++} and DeRF\cite{rebain2021derf} introduced optimized scene decompositions and novel optimization techniques, further accelerating training and inference. Additionally, techniques such as TensoRF~\cite{chen2022tensorf} and DVGO~\cite{sun2022directvoxelgridoptimization} leveraged tensor factorization and voxel grids to reduce memory requirements and computation time.
To address scalability, methods like Mip-NeRF 360~\cite{barron2021mipnerf}, Block-NeRF~\cite{tancik2022block}, and UrbanNeRF~\cite{rematas2022urban} introduced scene partitioning and multiscale representations for handling large and unbounded scenes. Meanwhile, approaches such as NeRF-W~\cite{martin2021nerf} disentangled appearance variations, and Ref-NeRF~\cite{verbin2022ref} extended NeRFs to handle reflective and refractive effects, advancing realism and robustness.
Additional advancements, such as Instant-NGP~\cite{M_ller_2022} and Zip-NeRF~\cite{barron2023zipnerfantialiasedgridbasedneural}, accelerated NeRF training by introducing a multi-resolution hash encoding and addressed aliasing issues inherent in grid-based models by incorporating anti-aliasing techniques, continuously pushing the boundaries of efficiency and quality in neural scene representation.

High-fidelity NeRF rendering requires dense ray sampling and extensive network evaluations, hindering real-time deployment. Its implicit scene representation also complicates integration with game engines and industrial pipelines, which prefer explicit geometry. These challenges drive the exploration of compact representations, explicit models, and novel optimization techniques to enable interactive, practical applications.

\begin{figure*}[htbp]
    \centering
    \includegraphics[width=0.95\linewidth]{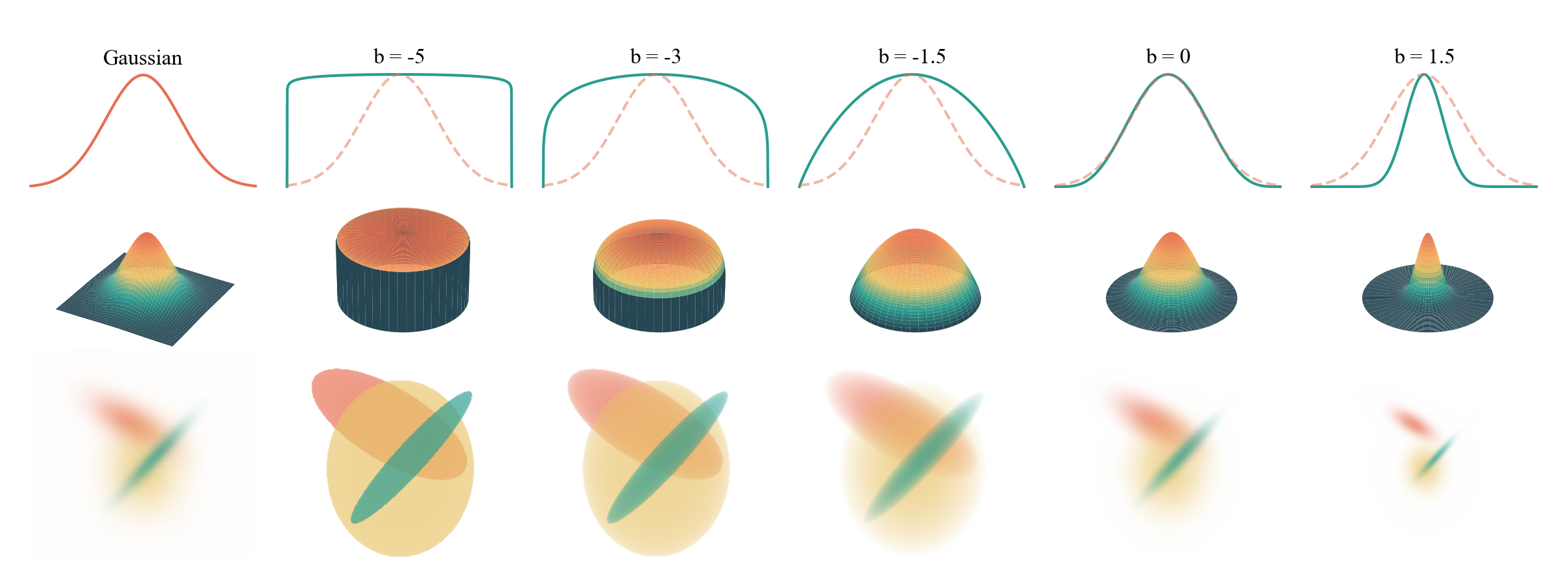}
    \caption{Unlike the fixed Gaussian Kernel, Deformable Beta Kernel adapts its shape to capture fine geometric and texture details. This figure shows how the kernel varies by different $b$ values as in \Cref{eq:beta}. Negative $b$ generates flatter surface but sharper cutoffs for learning solid support and sharp geometry, while positive $b$ generates sharper peaks for learning high frequency details. When $b=0$, the beta kernel is almost identical to Gaussian (domain scaled by $3\sigma$) } 
    \label{fig:betasplat}
\end{figure*}


\subsection{3D Gaussian Splatting}

3D Gaussian Splatting (3DGS)~\cite{3dgs} offers a compelling alternative to NeRF by representing scenes as collections of anisotropic Gaussian ellipsoids that capture local properties such as position, density, and view-dependent appearance. Combining point-based rendering\cite{zwicker2001ewa} with modern machine learning strategy, it enables flexible and efficient scene representation. However, despite its computational efficiency, 3DGS lags behind NeRF in photorealism, struggling with fine details, aliasing artifacts, and complex view-dependent effects.

Research has addressed these shortcomings across several fronts. Anti-aliasing methods such as Mip-Splatting~\cite{yu2023mipsplattingaliasfree3dgaussian}, multi-scale approaches~\cite{yan2024multiscale3dgaussiansplatting}, analytic approximations~\cite{liang2024analyticsplattingantialiased3dgaussian}, and scale-adaptive filtering~\cite{song2024sagsscaleadaptivegaussiansplatting} have been proposed to reduce jagged edges and shimmering, thereby smoothing the rendered output.
Deblurring techniques have also been explored to enhance clarity. Methods leveraging small MLPs~\cite{lee2024deblurring3dgaussiansplatting}, event camera data~\cite{peng2024bagsbluragnosticgaussian}, joint optimization of Gaussian parameters with camera trajectory recovery~\cite{zhao2024badgaussiansbundleadjusteddeblur,seiskari2024gaussiansplattingmoveblur,oh2024deblurgsgaussiansplattingcamera} have improved the sharpness of reconstructions from blurred inputs. These approaches integrate motion models and depth-aware filters into the 3DGS pipeline to refine both geometry and appearance.

Beyond addressing aliasing and blurring, splatting kernel modeling has emerged as an attractive direction for improving rendering quality and efficiency. 2DGS introduced a perspective-accurate 2D splatting pipeline that leverages ray–splat intersections and rasterization~\cite{2dgs}. Li \etal proposed 3D Half Gaussian Splatting, which optimizes the Gaussian representation by restricting it to half-space for better modeling of occlusions and interactions in complex 3D scenes~\cite{li20243dhgs3dhalfgaussiansplatting}. Hamdi \etal replaced Gaussians with a Generalized Exponential Function, capturing sharp edges with fewer primitives~\cite{hamdi2024gesgeneralizedexponentialsplatting}. Chen \etal developed Linear Kernel Splatting, swapping Gaussian kernels for linear ones to accelerate computation while preserving high-fidelity output~\cite{chen2024gaussiansfasthighfidelity3d}. Recent advances in splatting kernels include 3D Convex Splatting~\cite{held20243dconvexsplattingradiance}, which swaps 3D Gaussians for smooth, convex 3D primitives to better resolve edges and planes, and Deformable Radial Kernel Splatting~\cite{huang2025deformableradialkernelsplatting}, which enriches 2D Gaussians with learnable radial basis functions plus optimized ray–primitive intersection and culling for flexible, boundary-sharp shape modeling.
Despite advancements in splatting kernel designs, the absence of an aligned view-dependent color modeling and dedicated densification strategies for deformable kernels constrains further improvements in final rendering quality.

In the absence of a coherent view-dependent color framework, sophisticated appearance modeling becomes essential to elevate 3DGS rendering fidelity.
Xu \etal introduced Texture-GS, disentangling appearance from geometry by mapping 2D textures onto 3D surfaces via an MLP for high-fidelity editing~\cite{xu2024texturegsdisentanglinggeometrytexture}. Svitov \etal proposed BBSplat, using textured planar primitives with learnable RGB textures and alpha maps, achieving over 1200 FPS~\cite{svitov2024billboardsplattingbbsplatlearnable}. Weiss and Bradley developed Gaussian Billboards, unifying 2D Gaussian Splatting with UV-parameterized textures for expressive scene representations~\cite{weiss2024gaussianbillboardsexpressive2d}. Yang \etal introduced Spec-Gaussian, leveraging an MLP to model anisotropic spherical Gaussians for complex lighting~\cite{yang2024specgaussiananisotropicviewdependentappearance}.
While existing methods focus on either speed (via billboard textures) or fidelity (via MLPs, with slower rendering), we propose Spherical Beta, which explicitly models appearance by decoupling diffuse and specular components, inspired by the Phong Reflection Model~\cite{phong}.

Likewise, the lack of bespoke kernel densification strategies makes revisiting optimization and densification methods vital for unlocking further performance gains. Techniques such as Scaffold-GS~\cite{lu2023scaffoldgsstructured3dgaussians}, RadSplat~\cite{niemeyer2024radsplatradiancefieldinformedgaussian}, Pixel-GS~\cite{zhang2024pixelgsdensitycontrolpixelaware}, and AtomGS~\cite{liu2024atomgsatomizinggaussiansplatting} focus on efficient Gaussian placement and density control. Notably, 3DGS-MCMC~\cite{kheradmand20243dgaussiansplattingmarkov} introduces stochastic processes to refine the sampling and placement of Gaussians, enhancing both rendering quality and speed. While these strategies effectively capture scene details and balance rendering quality with computational cost, they are tailored specifically to Gaussian kernels. We therefore extend the principles of 3DGS-MCMC to propose a kernel-agnostic densification strategy that both stabilizes the optimization process and enforces consistent spatial distribution across arbitrary kernel types.


\section{Deformable Beta Splatting}
\label{sec:method}

The primary bottleneck in Gaussian Splatting–based approaches lies in their reliance on a fixed, smooth splatting kernel and low-order SH encoding, which inherently hampers their ability to capture flat surfaces and specular highlights. To overcome this limitation, we first derive a deformable Beta Kernel function. We then illustrate how this flexible kernel can be applied to both geometric splatting and color encoding, enabling the modeling of varying geometric complexities and lighting frequencies, which enhances the overall scene representation. Finally, we address the unique challenges introduced by employing a deformable kernel and detail how our optimization strategy effectively manages these issues.

\subsection{The Beta Kernel}
The Beta Kernel is derived from the Beta distribution \cite{johnson1995continuous}, defined as a power of the input $x$ and its reflection $(1-x)$:
\begin{equation}
    f(x;\alpha,\beta) = \frac{1}{\mathbf{B(\alpha, \beta)}} x^{\alpha - 1} (1-x)^{\beta - 1}, \quad x \in [0, 1],
\end{equation}
where $\mathbf{B}$ is the Beta Function for normalization, and $\alpha, \beta > 0$ are shape-controlling parameters. 

To obtain a bell-like shape that promotes smooth 3D view consistency, we fix $\alpha = 1$, focusing solely on the reflection component. By removing the normalization term, we simplify the Beta distribution to the Beta Kernel: 
\begin{equation}
    \mathcal{B}(x;\beta) = (1 - x)^{\beta}, \quad x \in [0, 1], \beta \in (0, \infty).
\end{equation}

The Beta Kernel is a power function over the closed domain $x \in [0, 1]$, inherently compact and bounded, with the parameter $\beta$ introducing adaptive shape control. It is worth noting that not every function maintains multi-view consistency or qualifies as a valid splatting kernel. We have employed the inverse Abel transform to prove that the Beta Kernel meets these criteria (please refer to \Cref{app: 3D Splatting Kernels} for details).

However, direct optimization of $\beta$ can bias the kernel toward low-frequency representations because as $\beta$ approaches zero, the Beta Kernel emphasizes lower frequencies, while high-frequency shapes require $\beta \to \infty$. To allow for unbiased optimization while maintaining the proper range of $\beta$, we reparameterize $\beta$ using an exponential activation: $\beta(b)=e^b$. Moreover, we want the Beta Kernel to initially resemble a Gaussian-like shape and then gradually adapt to learn different geometry forms. To achieve this, we compute a constant $c=4$ such that $\int_0^1(1-x)^{ce^b}dx \approx \int_0^1 e^{-\frac{9x}{2}}dx$ when $b=0$. This ensures that the kernel starts with a Gaussian-like function when $b$ is zero initialized.

As a result, we parameterize the Beta Kernel as:
\begin{equation}
\label{eq:beta}
    \mathcal{B}(x; b) = (1 - x)^{\beta(b)}, \quad \beta(b) = 4e^{b}, \quad x \in [0, 1], \quad b \in \mathbb{R}.
\end{equation}

This parameterization ensures that the Beta Kernel begins with a Gaussian-like shape when $b$ is initialized to zero and can adapt its shape during optimization. It eliminates the need for manual fine-tuning of initial $\beta$ values, effectively making the Beta Kernel a more flexible superset of the Gaussian Kernel.

\subsection{3D Ellipsoidal Beta Primitive}
We define a 3D Ellipsoidal Beta Primitive as a set of parameters:
\begin{equation}
    B = \{\bm{\mu}, o, \bm{q}, \bm{s}, b, \bm{f}\}
\end{equation}
where the geometry parameters $\bm{\mu}, o, \bm{q}, \text{and }\bm{s}$ mirror those in 3DGS. Specifically, $\bm{\mu} \in \mathbb{R}^3$ denotes the center of the primitive in 3D space, $o \in [0, 1]$ represents opacity, $\bm{q} \in [-1, 1]^4$ defines the rotation through a quaternion, and $\bm{s} \in [0, \infty)^3$ determines the scale of the ellipsoidal shape.

The key innovation in our approach is the introduction of the parameter $b \in \mathbb{R}$, which controls the shape of the Beta Kernel.  As illustrated in \Cref{fig:betasplat}, varying $b$ deforms the kernel shape, with $b=0$ yielding a function nearly identical to a Gaussian. This additional parameter provides adaptive shape control with minimal overhead, allowing the kernel to dynamically adjust from a Gaussian-like function to forms that can capture sharper edges and flat surfaces as optimization progresses. 

For appearance modeling, each primitive carries a feature vector $f\in \mathbb{R}^d$ that encodes view-dependent color information. This vector is crucial for representing complex lighting interactions and will be discussed in detail in the next subsection.

The rendering process for a pixel $\bm{x} \in \mathbb{R}^2$ unfolds as follows: each 3D Ellipsoidal Beta Primitive is projected into the 2D image plane using the viewing transformation $\bm{W}$. This results in a 2D projected center $\bm{\mu}^\prime \in \mathbb{R}^2$ and a corresponding 2D covariance matrix $\bm{\Sigma}^\prime = \bm{JW\Sigma W^\top J^\top}$, where $\Sigma=RSS^\top R^\top$ with $R$ and $S$ derived from the quaternion and scale, respectively. We then compute the distance between the pixel and each primitive center:
    \begin{equation}
        r_i(\bm{x}) = \sqrt{(\bm{x} - \bm{\mu}_i^\prime)^\top \bm{\Sigma}_i^{\prime^{-1}} (\bm{x} - \bm{\mu}_i^\prime)}.
    \end{equation}
After sorting overlapping primitives by depth relative to the camera, denoted by the ordered set $\mathcal{N} = \{B_1, \dots, B_N\}$, we composite their contributions to the final pixel color $\bm{C}(\bm{x})$ using the Beta Kernel:
\begin{equation}
    \bm{C}(\bm{x}) = \sum_{i=1}^N c_i o_i \mathcal{B}(r_i(\bm{x})^2; b_i) \prod_{j=1}^{i-1} \left(1 - o_j \mathcal{B}(r_j(\bm{x})^2; b_j)\right)
\end{equation}

\begin{figure}
    \centering
    \includegraphics[width=\linewidth]{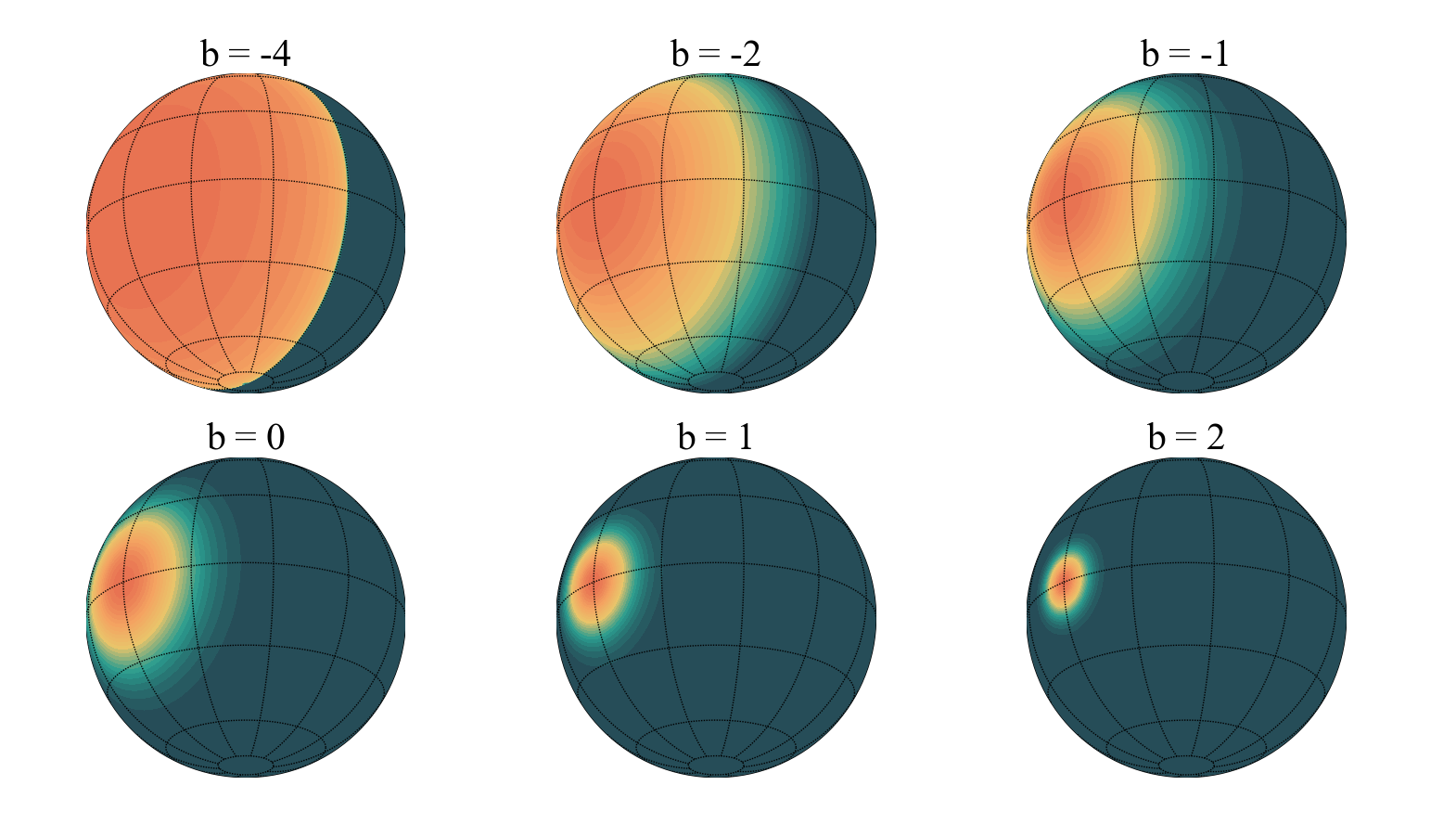}
    \caption{Learnable Spherical Beta can effectively capture specular highlights with varying sharpness. Smaller $b$ values result in broader reflections, while higher $b$ values correspond to sharper specular highlights.}
    \label{fig:SB}
\end{figure}

\subsection{Spherical Beta}
3DGS employs Spherical Harmonics (SH) for view-dependent color encoding. For a SH of degree $N$, the feature dimension scales as $3(N+1)^2$, resulting in a quadratic increase in parameters. To achieve real-time performance in practical applications, 3DGS is constrained to using low-order SH (\eg,$N=3$), which only provide smooth view-dependent colors and struggle to model sharp specular highlights effectively.
To address this, we introduce Spherical Beta (SB) function, which is inspired by the Phong Reflection Model~\cite{phong}:
\begin{equation}
    c(\hat{V}) = A_m + \sum_{m \in \mathcal{M}} \left[D_m + (\hat{R}_m \cdot \hat{V})^{\alpha_m} c_m\right],
\end{equation}
where $A_m$ is the ambient term, $D_m$ and $(\hat{R}_m \cdot \hat{V})^{\alpha_m} c_m$ models the diffuse and specular components for light source $m$. Here, $\hat{R}_m$ is the normalized reflection direction, $\hat{V}$ is the normalized view direction, $\alpha_m$ is the shininess coefficient controlling specularity, and $c_m$ is the color of light source.
Note that the reflection vector $\hat{R}_m$ is derived from the surface normal and corresponding light source directions.
In contrast, Spherical Beta merges ambient and diffuse into a single base color, then directly models specular lobes via learnable bounded Beta kernels.
\begin{equation}
    c(\hat{V}) = c_0 +  \sum_{m \in \mathcal{M}}\mathcal{B}(1-\hat{R}_m \cdot \hat{V};b_m) c_m
\end{equation}
where $c_0$ represents the diffuse or base color for the Beta primitive. For each 3D Beta primitive, we introduce $M=|\mathcal{M}|$ SB lobes modeling outgoing reflected radiance. For each SB lobe $m\in\mathcal{M}$, it is paramertized by reflection direction $\hat{R}_m$, color $c_m$, and shininess $b_m$. We can observe that its feature dimension scales as $3+6M$, growing linearly with the number of reflection lobes and making it more efficient than SH.

Spherical Beta shares similarities with Spherical Gaussian in modeling specularities but offers improvements. Unlike Spherical Gaussian, which have unbounded support and require truncation—leading to radiance discontinuities when view and reflection directions are orthogonal—the Beta Kernel is inherently bounded. Its input range is confined to $[0,1]$, ensuring that radiance naturally drops to zero when $\hat{R}_m \cdot \hat{V}=0$. This eliminates artifacts associated with cutoff truncation and maintains continuity in radiance.

\Cref{fig:SB} illustrates the specular highlights and corresponding reflection lobes for different $b_m$ values, demonstrating the capability of Spherical Beta to model both high and low frequency specularities. By integrating bounded support and adaptive shininess, Spherical Beta enhances the ability to model complex lighting interactions and specular effects with fewer parameters, thereby improving both visual fidelity and computational efficiency.

\begin{figure*}[htbp]
    \centering
    \includegraphics[width=0.9\linewidth]{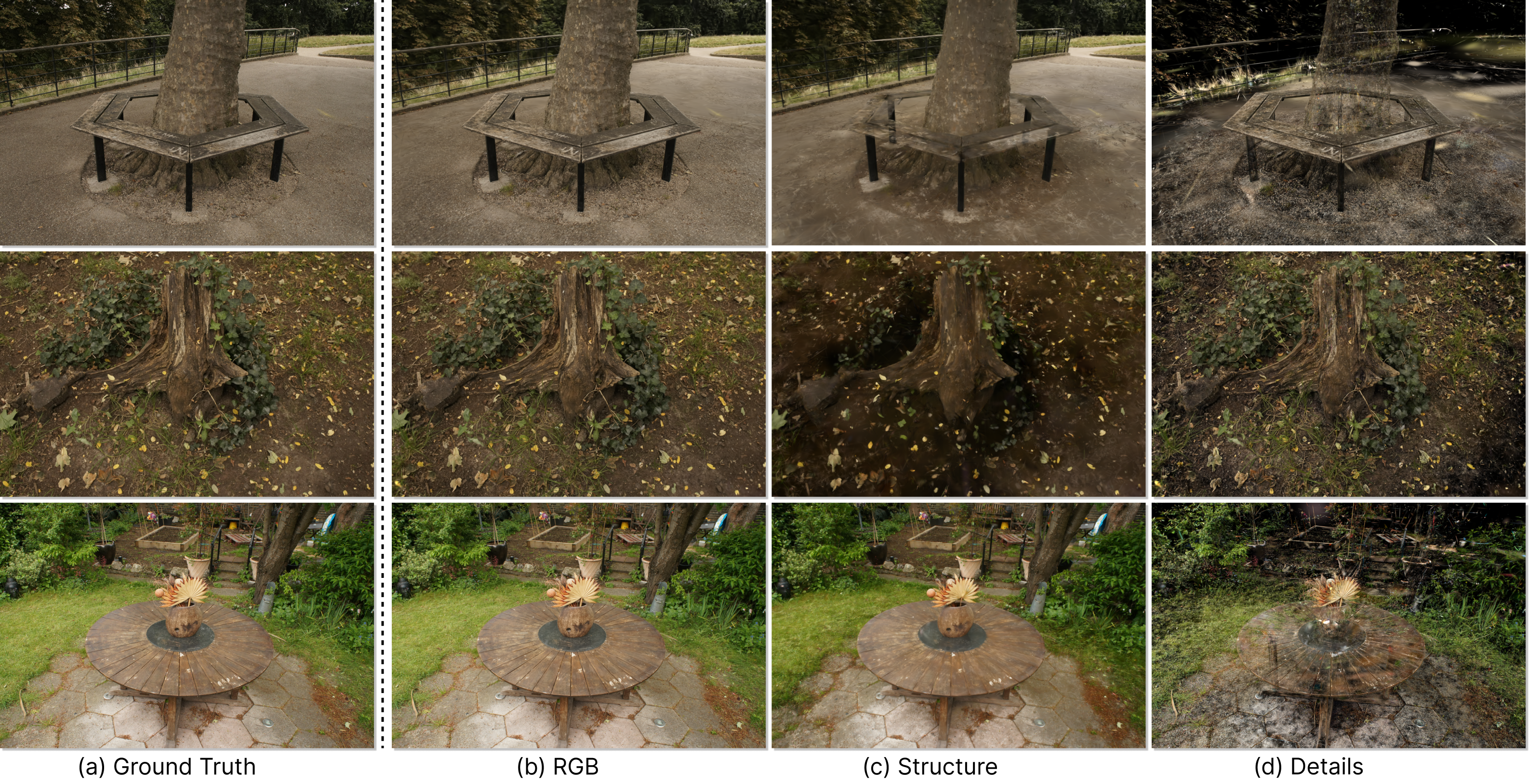}
    \caption{Geometry Decomposition: our designed Beta Kernel provides the capability to decompose scene geometry into fundamental structures and intricate details, such as high-frequency textures and fine surface variations.}
    \label{fig:lod}
\end{figure*}

\begin{figure*}[htbp]
    \centering
    \includegraphics[width=0.9\linewidth]{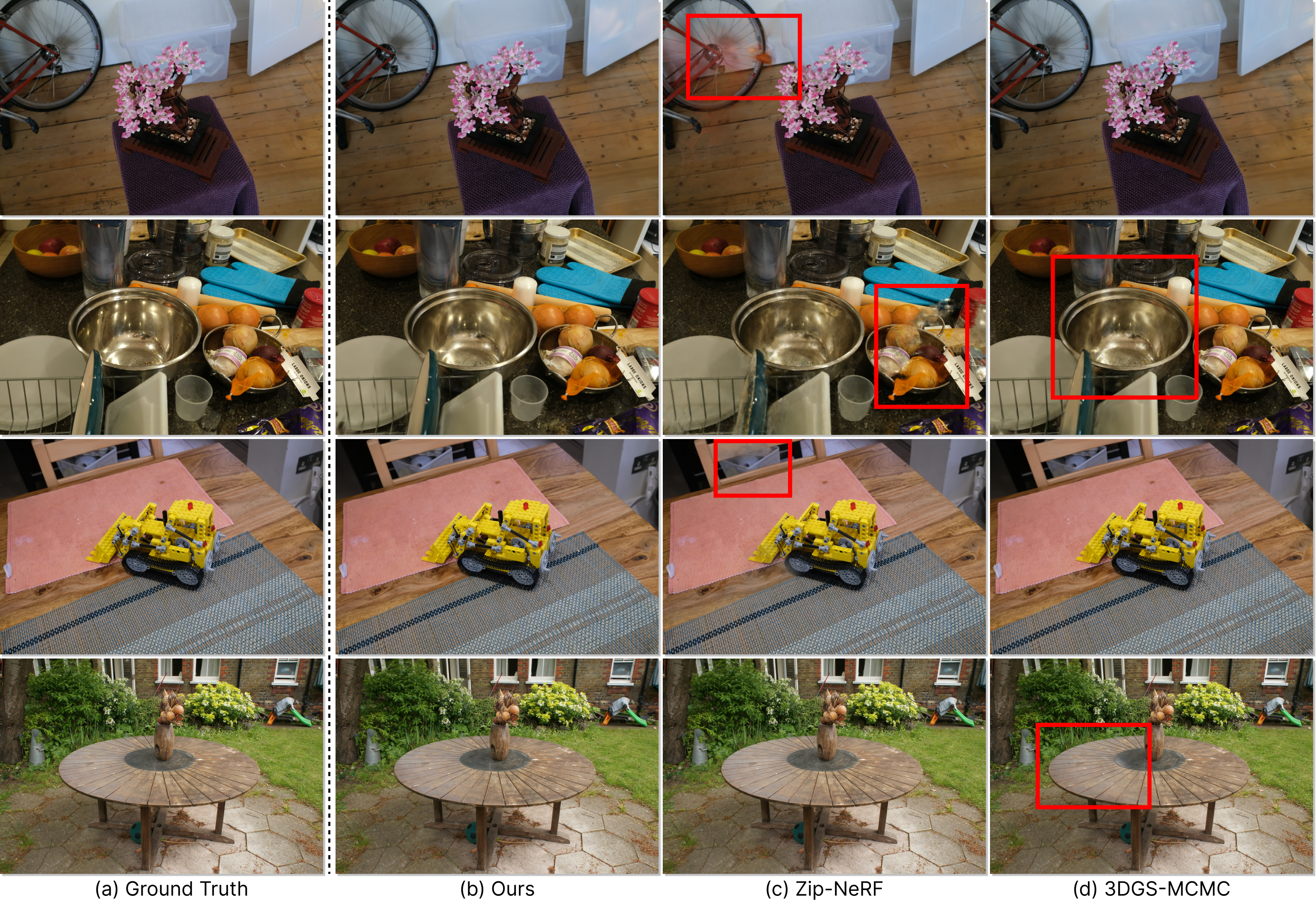}
    \vspace{-0.8em} 
    \caption{Qualitative Comparison: Leveraging our Deformable Beta Kernel, Spherical Beta color encoding, and Kernel-Agnostic MCMC optimization process, the Deformable Beta Splatting framework qualitatively outperforms the implicit SOTA method (Zip-NeRF) and the explicit SOTA method (3DGS-MCMC).}
    \label{fig:qualitative}
\end{figure*}

\subsection{Kernel-Agnostic Markov Chain Monte Carlo}
One of the key innovations of 3DGS is its optimization with densification strategy, which utilizes heuristic gradient thresholds to clone, split, and prune primitives. Additionally, 3DGS-MCMC redefines Gaussian Splatting optimization as a Markov Chain Monte Carlo (MCMC) process by treating each primitive's opacity as a probability. This approach identifies "dead" primitives—those with opacity below a pruning threshold—and respawns them by relocating to live Gaussians through multinomial sampling based on their opacity values. By adjusting both opacity and scale, this method preserves the underlying probability distribution during densification, thereby stabilizing the process. Moreover, the introduction of positional noise encourages exploration and prevents overfitting.

However, these heuristic strategies and scale adjustments are inherently tied to the properties of the Gaussian function, making them difficult to directly apply to arbitrary kernels.

To address this limitation, we present Kernel-Agnostic MCMC, building upon the framework established by 3DGS-MCMC~\cite{kheradmand20243dgaussiansplattingmarkov}. We first detail the optimization process and then revisit distribution-preserved densification.

\paragraph{Optimization}

To encourage exploration and prevent overfitting, we adopt a noise term \( \epsilon \) that perturbs the positions of primitives during optimization.
The primitive positions \( \bm{\mu} \) are then updated as
\begin{equation}
    \bm{\mu} \leftarrow \bm{\mu} - \lambda_{\text{lr}} \cdot \nabla_{\bm{\mu}} \mathcal{L} + \lambda_\epsilon \cdot \epsilon,  \quad   \epsilon = \lambda_{\text{lr}} \cdot \mathcal{B}(o_i; b^\prime) \cdot \Sigma_\eta.
\end{equation}
Note that we replace the logit noise function from 3DGS-MCMC with a Beta Kernel function for a more compact and well-defined noise function.
Here, \( \lambda_\epsilon \) controls the impact of the noise term, ensuring that the primitive is not only optimized through backpropagation of the loss function but also explores alternative solutions for better representation. \( \lambda_{\text{lr}} \) represents the learning rate, and \( \Sigma_\eta \) is the sampled noise. \( \mathcal{B}(o_i; b^\prime) \) is the Beta Kernel component, which scales the noise based on the opacity of each primitive, with \( b^\prime = \ln(25) \) to resemble the fall-off behavior of the original logit function.

To train the Beta Primitives, we employ a composite loss function:
\begin{equation}
       \mathcal{L} = (1 - \lambda_{\text{SSIM}}) \mathcal{L}_1 + \lambda_{\text{SSIM}} \mathcal{L}_{\text{SSIM}} 
    + \lambda_o \sum_i |o_i| 
    + \lambda_\Sigma \sum_i \sum_j |\sqrt{\text{eig}_j(\Sigma_i)}|, 
\end{equation}
where \( \mathcal{L}_1 \) is the pixel-wise L1 loss that enforces accurate color reproduction in the rendered image, and \( \mathcal{L}_{\text{SSIM}} \) is the Structural Similarity Index Measure (SSIM) loss that promotes structural consistency across the scene. The hyperparameters \( \lambda_{\text{SSIM}}, \lambda_o, \) and \( \lambda_\Sigma \) balance the contributions of each loss term. 

The term \( \lambda_o \sum_i |o_i| \) acts as an opacity regularizer to ensure the kernel-agnostic densification and encourage primitive to explore. The eigenvalues \( \text{eig}_j(\Sigma_i) \) of the covariance matrix \( \Sigma_i \) for the \( i \)-th primitive are used in the regularization term \( \lambda_\Sigma \sum_i \sum_j |\sqrt{\text{eig}_j(\Sigma_i)}| \), which together with opacity regularizer encourage primitives to disappear and respawn.

\paragraph{Densification}
Here, we mathematically demonstrate that regularizing opacity and adjusting it alone are both effective and valid, regardless of the number of clones or the chosen splatting kernel.

Assume a primitive defined by a splatting kernel $f(x)$ with opacity $o$ and it will be copied $N$ times, ~\cite{bulò2024revisingdensificationgaussiansplatting, kheradmand20243dgaussiansplattingmarkov} established that
\begin{equation}
    o' = 1 - \sqrt[N]{1 - o},
\end{equation}

Incorporating $f(x)$ complicates preservation beyond the primitive center, especially at boundaries. Prior approaches like 3DGS-MCMC handled boundary preservation via scale adjustments tailored to Gaussian kernels, which are complex and not directly applicable to deformable kernels.

However, we find that given a regularized small $o$ value, we approximate $o'$ using a Taylor expansion $\sqrt[N]{1 - o} \approx 1 - \frac{o}{N}
$, such that 
\begin{equation}
    o' = 1 - \sqrt[N]{1 - o} \approx \frac{o}{N}
\end{equation}

Now consider the densified distribution:
\begin{equation}
    1 - \Big(1 - o' \cdot f(x)\Big)^N\approx 1 - \Big(1 - \frac{o}{N} f(x)\Big)^N.
\end{equation}

Since $\frac{o}{N} f(x)$ is also small, applying the binomial approximation:
\begin{equation}
\Big(1 - \frac{o}{N} f(x)\Big)^N \approx 1 - N \cdot \frac{o}{N} f(x) + \mathcal{O}(o^2) = 1 - o \, f(x) + \mathcal{O}(o^2),
\end{equation}
 we find:
\begin{equation}
1 - \big[1 - o \, f(x) + \mathcal{O}(o^2)\big] = o \, f(x) + \mathcal{O}(o^2),
\end{equation}
which matches the original distribution $o \cdot f(x)$.

The error term $\mathcal{O}(o^2)$ becomes negligible as we employ an opacity regularizer to maintain a small $o$. Therefore, the difference between original distribution and densified distribution diminishes by regularizing opacity regardless of copy times $N$ and splatting kernel $f(x)$. Detailed proof can be found in \Cref{app:densification}.

\begin{figure*}
    \centering
    \includegraphics[width=0.9\linewidth]{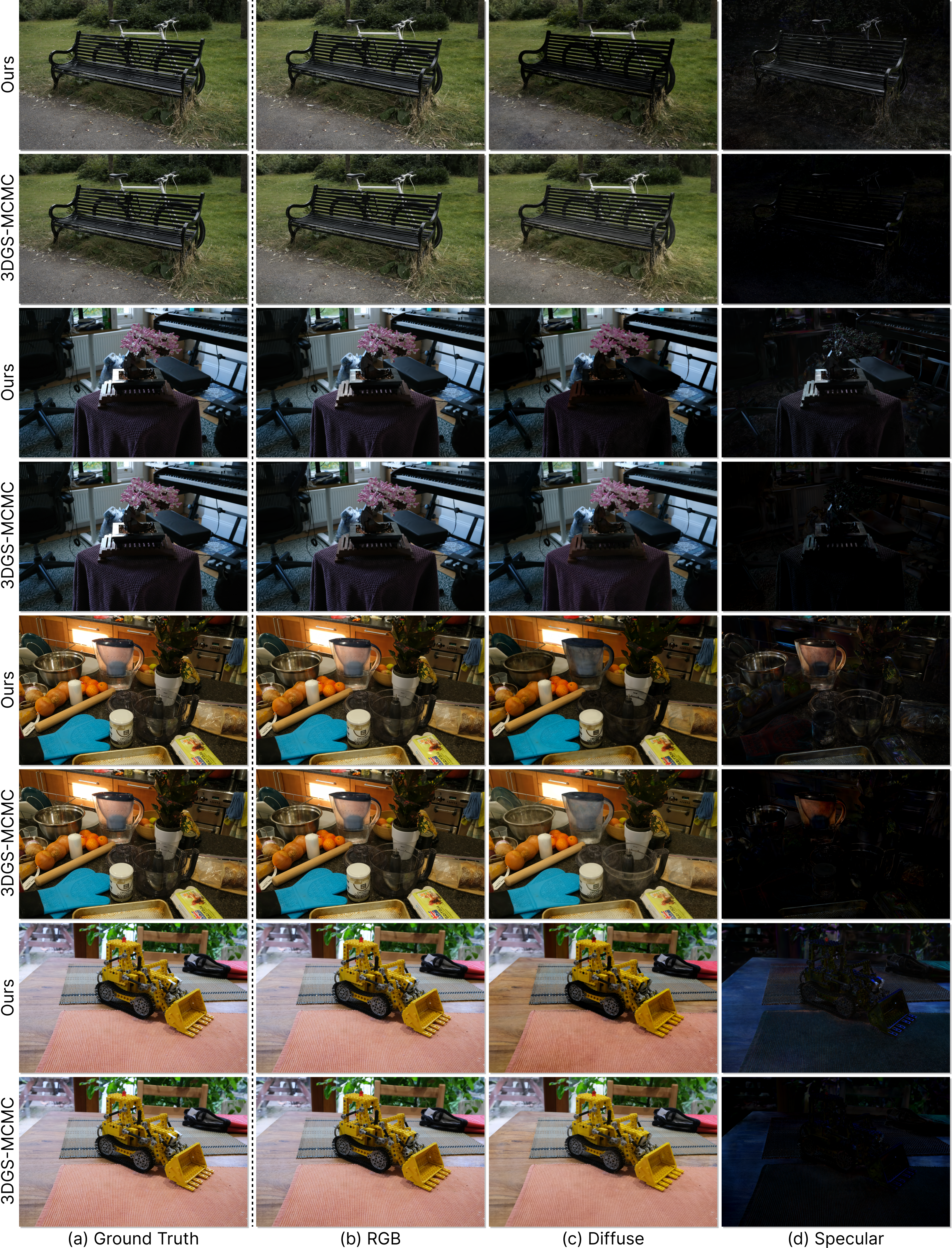}
    \caption{Light Decomposition: our Spherical Beta color encoding can effectively decompose diffuse and specular components compared to 3DGS-based approaches using low-order Spherical Harmonics, as illustrated in this figure.}
    \label{fig:light}
\end{figure*}

\section{Experiments}
\label{sec:experiments}

\begin{table*}[htbp]
    \centering
    \resizebox{0.95\linewidth}{!}{%
    \begin{tabular}{cc|ccc|ccc|ccc|ccc}
        \hline
        & \multirow{2}{*}{Methods} 
        & \multicolumn{3}{c|}{Mip-NeRF360} 
        & \multicolumn{3}{c|}{Tanks\&Temples} 
        & \multicolumn{3}{c|}{Deep Blending}
        & \multicolumn{3}{c}{NeRF Synthetic}\\
         & 
         & PSNR$\uparrow$ & SSIM$\uparrow$ & LPIPS$\downarrow$
         & PSNR$\uparrow$ & SSIM$\uparrow$ & LPIPS$\downarrow$
         & PSNR$\uparrow$ & SSIM$\uparrow$ & LPIPS$\downarrow$
         & PSNR$\uparrow$ & SSIM$\uparrow$ & LPIPS$\downarrow$ \\
        \hline
        \parbox[t]{2mm}{\multirow{3}{*}{\rotatebox[origin=c]{90}{implicit}}}
         & Instant-NGP~\cite{M_ller_2022}
           & 25.51 & 0.684 & 0.398
           & 21.72 & 0.722 & 0.330
           & 23.62 & 0.797 & 0.423 
           &33.18 & 0.959&0.055\\
         & Mip-NeRF360~\cite{barron2021mipnerf}
           & 27.69 & 0.792 & 0.237
           & 22.22 & 0.758 & 0.257
           & 29.40 & 0.900 & 0.245
           &33.25&0.962&0.039\\
         & Zip-NeRF~\cite{barron2023zipnerfantialiasedgridbasedneural}
           & \cellcolor{yellow!25}28.54 & 0.828 & \cellcolor{yellow!25}0.189 
           & - & - & - 
           & - & - & - 
           &33.10&\cellcolor{yellow!25}0.971&\cellcolor{yellow!25}0.031\\
        \hdashline
        \parbox[t]{2mm}{\multirow{7}{*}{\rotatebox[origin=c]{90}{explicit}}}
         & 3DGS~\cite{3dgs} 
           & 27.20 & 0.815 & 0.214 
           & 23.15 & 0.840 & \cellcolor{yellow!25}0.183 
           & 29.41 & \cellcolor{yellow!25}0.903 & \cellcolor{yellow!25}0.243 
           &33.31&0.969&0.037\\
         & GES~\cite{hamdi2024gesgeneralizedexponentialsplatting}&26.91&0.794&0.250&23.35&0.836&0.198&\cellcolor{yellow!25}29.68&0.901&0.252& - & - & -  \\
         & 2DGS~\cite{2dgs} 
           & 27.04 & 0.805 & 0.297 
           & - & - & - 
           & - & - & - 
           &33.07&-&-\\
         & Mip-Splatting~\cite{yu2023mipsplattingaliasfree3dgaussian} 
           & 27.79 & 0.827 & 0.203 
           & - & - & - 
           & - & - & - 
           &33.33&0.969&0.039\\
         & 3DGS-MCMC~\cite{kheradmand20243dgaussiansplattingmarkov} 
           & 28.29 & \cellcolor{yellow!25}0.840 & 0.210 
           & \cellcolor{yellow!25}24.29 & \cellcolor{yellow!25}0.860 & 0.190 
           & 29.67 & 0.895 & 0.320 
           &\cellcolor{yellow!25}33.80&0.970&0.040\\
         & Ours (30k) 
           & \cellcolor{orange!25}28.60 & \cellcolor{orange!25}0.844 & \cellcolor{orange!25}0.182
           & \cellcolor{orange!25}24.79 & \cellcolor{orange!25}0.868 & \cellcolor{orange!25}0.148
           & \cellcolor{orange!25}30.10 & \cellcolor{orange!25}0.910 & \cellcolor{orange!25}0.240 &
           \cellcolor{orange!25}34.64 & \cellcolor{orange!25}0.973 &\cellcolor{orange!25}0.028\\
         & Ours (full) 
           & \cellcolor{red!25}28.75 & \cellcolor{red!25}0.845 & \cellcolor{red!25}0.179 
           & \cellcolor{red!25}24.85 & \cellcolor{red!25}0.870 & \cellcolor{red!25}0.140 
           & \cellcolor{red!25}30.12 & \cellcolor{red!25}0.914 & \cellcolor{red!25}0.236 &
           \cellcolor{red!25} 34.66 & \cellcolor{red!25}0.973 & \cellcolor{red!25}0.028\\
        \hline
    \end{tabular}}
    \caption{Quantitative evaluation was conducted on the Mip-NeRF360 \cite{barron2021mipnerf}, Tanks\&Temples \cite{Knapitsch2017TanksAndTemples}, and Deep Blending \cite{DeepBlending} datasets. We adopt the original number from corresponding papers when available. A dash (``-'') indicates results that were not reported. }
    \label{tab:metrics_nonerfsynthetic}
\end{table*}

We evaluate Deformable Beta Splatting (DBS) on a diverse set of real-world and synthetic datasets to demonstrate its superior visual quality and rendering efficiency compared to existing state-of-the-art methods. Additionally, DBS effectively decouples scene geometry and lighting information, leveraging the learnable frequency characteristics of the Beta Kernel to enhance scene representation.

\subsection{Experimental Setup}

\paragraph{Implementations.} Our implementation builds upon the 3DGS-MCMC framework \cite{kheradmand20243dgaussiansplattingmarkov}, incorporating modifications to the GSplat library \cite{ye2024gsplatopensourcelibrarygaussian}. Specifically, we integrate the Beta Kernel and Spherical Beta (SB) functions, which are fully explicit and differentiable, implemented in CUDA for optimized performance.
By default, we utilize two light sources in Spherical Beta component per primitive (i.e., $sb=2$), unless otherwise specified. To ensure that the Beta Kernel initially resembles the Gaussian Kernel, we set $b$ to zero at initialization. 
We observe that different scenes exhibit varying convergence speeds when training the model for the default 30,000 iterations. To accommodate these variations, we have integrated a flexible stopping strategy with patience of 10k iterations into the framework, allowing for two training modes: (Ours, 30K) and (Ours, Full). Detailed training times and iteration counts for each scene are provided in \Cref{app:additional}. For additional hyperparameters and configuration settings, please refer to \Cref{app:hyperparameters}.

\subsection{Results and Comparisons}

\paragraph{Radiance Field Reconstruction.} We evaluate DBS using the standard 3DGS evaluation pipeline across several widely recognized datasets, including Mip-NeRF 360 \cite{barron2021mipnerf}, Tanks and Temples \cite{Knapitsch2017TanksAndTemples}, Deep Blending \cite{DeepBlending}, and the NeRF Synthetic dataset \cite{mildenhall2021nerf}. Performance is measured using standard metrics: Peak Signal-to-Noise Ratio (PSNR), Structural Similarity Index Measure (SSIM), and Learned Perceptual Image Patch Similarity (LPIPS).
For the Mip-NeRF 360, Tanks and Temples, and Deep Blending datasets, we utilize the Colmap Structure-from-Motion (SfM) point cloud provided by the 3DGS \cite{3dgs} repository for initialization. In contrast, for the NeRF Synthetic dataset, primitives are randomly initialized to simulate diverse starting conditions. We benchmark DBS against prior works, including the state-of-the-art implicit method (ZipNeRF \cite{barron2023zipnerfantialiasedgridbasedneural}) and explicit method (3DGS-MCMC \cite{kheradmand20243dgaussiansplattingmarkov}). The quantitative results of comparisons are presented in \Cref{tab:metrics_nonerfsynthetic}.

Beta Splatting demonstrates superior performance for radiance field reconstruction across all metrics over almost all datasets compared to prior methods. Both the 30K and Full training modes consistently outperform previous implicit and explicit state-of-the-art methods. See \Cref{app:additional} for detailed per-scene analysis.
Qualitative visual comparisons in \Cref{fig:qualitative} reveal that while Zip-NeRF tends to exhibit floater artifacts, and 3DGS-MCMC struggles with rendering sharp specularities, DBS maintains high fidelity in both smooth and highly detailed regions, effectively capturing intricate lighting interactions and geometric details.

\paragraph{Efficiency Analysis.}
\begin{table}[htbp]
    \centering
     \resizebox{0.9\linewidth}{!}{\begin{tabular}{c|ccc}
    \hline
        
         \multirow{2}{*}{Method}& \multicolumn{3}{c}{Mip-NeRF 360}\\
         &$\text{Training Time}^\downarrow$ &$\text{Storage Mem}^\downarrow$& $\text{Rendering FPS}^\uparrow$ \\
         \hline
         Zip-NeRF & 5h29m & \cellcolor{orange!25}569.7MB & 0.184\\
         3DGS & \cellcolor{orange!25}21m & 797.06MB & \cellcolor{yellow!25}116.15 \\
         3DGS-MCMC & 31m & \cellcolor{yellow!25}733.19MB & 82.46\\
         (Ours,30k,sb=2)& \cellcolor{red!25}16m & \cellcolor{red!25}356.04MB & \cellcolor{red!25}123.11 \\
         (Ours,full,sb=2)& \cellcolor{yellow!25}22m & \cellcolor{red!25}356.04MB & \cellcolor{orange!25}123.09\\
        \hline
    \end{tabular}}
    \caption{Quantitative evaluation of training and rendering efficiency.}
    \label{tab:efficiency}
\end{table}
To evaluate efficiency, we compare our method with Zip-NeRF \cite{barron2023zipnerfantialiasedgridbasedneural} and 3DGS-MCMC \cite{kheradmand20243dgaussiansplattingmarkov} in terms of training efficiency, rendering efficiency, and model size on a single NVIDIA RTX 6000 Ada Generation GPU.
The summarized efficiency metrics are presented in \Cref{tab:efficiency}, with additional per-scene details available in \Cref{app:additional}.
Our findings indicate that DBS not only achieves better reconstruction quality but also renders 1.5x faster than 3DGS-MCMC \cite{kheradmand20243dgaussiansplattingmarkov} under the same number of primitives. When sb=2, our method uses less than half storage memory compared to 3DGS-based methods while achieving the state-of-the-art in all metrics. 
Additional details on applying compression methods to further reduce storage requirements are provided in \Cref{app:compression}.
These results highlight effectiveness of DBS in delivering high-quality renderings with enhanced computational efficiency.

\paragraph{Ablation Study.} 
\begin{table}[htbp]
    \centering
     \resizebox{0.9\linewidth}{!}{\begin{tabular}{c|cc|cc}
    \hline
        
         \multirow{2}{*}{Methods}& \multicolumn{2}{c|}{Tanks\&Temples} & \multicolumn{2}{c}{Deep Blending} \\
        &$\text{PSNR}^\uparrow$ &$\text{Mem (MB)}^\downarrow$&$\text{PSNR}^\uparrow$ &$\text{Mem (MB)}^\downarrow$\\
        \hline
        3DGS, w/ MCMC, sh=3&24.29&437.63&29.67&750.23\\
        DBS, w/ MCMC, sh=3&24.45&445.05&29.88&762.94\\
        DBS, w/ Kernel-Agnostic MCMC, sh=3&24.57&445.05&30.03&762.94\\
        DBS, w/ Kernel-Agnostic MCMC, sb=1&24.53&\cellcolor{red!25}155.77&30.04&\cellcolor{red!25}267.03\\
        DBS, w/ Kernel-Agnostic MCMC, sb=2&24.79&\cellcolor{orange!25}200.27&\cellcolor{orange!25}30.10&\cellcolor{orange!25}343.32\\
        DBS, w/ Kernel-Agnostic MCMC, sb=4&\cellcolor{orange!25}24.81&\cellcolor{yellow!25}289.28&\cellcolor{red!25}30.12&\cellcolor{yellow!25}495.91\\
        DBS, w/ Kernel-Agnostic MCMC, sb=6&\cellcolor{yellow!25}24.80&378.29&\cellcolor{yellow!25}30.08&648.50\\
        \hdashline
        DBS (full)&\cellcolor{red!25}24.85&\cellcolor{orange!25}200.27&\cellcolor{red!25}30.12&\cellcolor{orange!25}343.32\\
    \hline
    \end{tabular}}
    \caption{Ablation Studies on Tank\&Temples and Deep Blending datasets.}
    \label{tab:ablation}
\end{table}
\Cref{tab:ablation} compares PSNR and memory usage on the Tanks \& Temples and Deep Blending datasets across several DBS configurations: moving from 3DGS-MCMC to a DBS-MCMC raises PSNR from 24.29 dB to 24.45 dB (Tanks \& Temples) and from 29.67 dB to 29.88 dB (Deep Blending); switching to Kernel-Agnostic MCMC further boosts PSNR to 24.57 dB and 30.03 dB with no added memory; varying the number of spherical-beta light sources (sb) reveals a trade-off: sb = 1 uses only 155.8 MB but yields 24.53 dB/30.04 dB, sb = 2 hits a sweet spot at 200.3 MB for 24.79 dB/30.10 dB, while sb = 4 or 6 slightly increase PSNR (up to 24.81 dB/30.12 dB) at much higher memory (289–378 MB); finally, adding our flexible stopping strategy to the sb = 2 setup (“DBS (full)”) delivers a final bump to 24.85 dB and 30.12 dB.

\subsection{Beta Decomposition}
Leveraging the deformable nature of the Beta Kernel, our framework enables decomposition of both geometry and lighting post-training. This capability arises from the dual role of the Beta Kernel in encoding volumetric splatting and view-dependent color information.

\paragraph{Geometry Decomposition.}
The Beta Kernel's parameter $b$ intrinsically controls the geometry frequency representation of each primitive. Lower $b$ values correspond to lower-frequency primitives, which predominantly capture the fundamental geometric structures of the scene. Conversely, higher $b$ values encapsulate higher-frequency details, such as textures and fine surface variations.

By setting a mask based on the $b$ parameter, we can selectively isolate primitives based on their frequency contributions. We can also do a simple binary split based on a threshold (\eg the mean of $b$ values across all primitives). Primitives with 
$b$ below the threshold primarily represent the scene's solid geometry,  with primitives of $b$ above the threshold provides high-frequency texture details.

\Cref{fig:lod} illustrates how Beta Kernels separate geometry from texture on real-world datasets. In the "treehill" scene, the bark texture is represented using primitives with beta parameter $b$. Similarly, in the "garden" scene, detailed textures on the tabletop are removed, preserving the overall table geometry.

\paragraph{Lighting Decomposition}
Since Beta Splatting integrates a learnable Phong Reflection Model \cite{phong}, we can extract diffuse component through taking only the base color $c_0$. Conversely, only enabling specular component gives the view-dependent specular lighting effects. For comparison, we only enable zero degree of SH for diffuse color while non zero degree for specular effects.

The results in \Cref{fig:light} illustrate the effectiveness of this approach. In the "bicycle" scene, the black bench is separated from the white highlights, a result unattainable with Gaussian Splatting. Similarly, in the "bonsai" scene, Beta Splatting successfully reconstructs the bright highlight component of the bonsai container, while Gaussian Splatting fails to capture such extreme view-dependent details. 


\section{Conclusion}

We presented Deformable Beta Splatting (DBS), a novel approach that advances real-time radiance field rendering through three key innovations: Deformable Beta Kernels for adaptive geometry representation, Spherical Betas for efficient view-dependent color encoding, and Kernel-Agnostic MCMC that enhances optimization stability and efficiency by relying solely on regularized opacity. These innovations allow DBS to achieve superior visual quality with less memory and computing resources than prior methods.

\paragraph{Limitations} As our framework is rasterization-based, it occasionally produces popping artifacts due to depth approximation inaccuracies during the sorting process. While adaptive, Spherical Beta Functions struggle to model mirror-like reflections and anisotropic specular highlights effectively. In addition, the frustum model causes Beta Kernels to optimize flat geometries for distant backgrounds, which impacts the overall beta geometry distribution. These limitations offer opportunities for further refinement and enhancement of the framework.
\label{sec:conclusion}

\begin{acks}
The authors would like to thank our primary sponsors of this research: Mr. Clayton Burford of the Battlespace Content Creation (BCC) team at Simulation and Training Technology Center (STTC). This work is supported by University Affiliated Research Center (UARC) award W911NF-14-D-0005, a Toyota Research Institute University 2.0 program, a gift fund from Dolby, and a gift fund from Google DeepMind. Statements and opinions expressed and content included do not necessarily reflect the position or the policy of the Government, and no official endorsement should be inferred.

\end{acks}
\bibliographystyle{ACM-Reference-Format}
\bibliography{sample-bibliography}


\clearpage

\maketitlesupplementary
\appendix

\section{3D Splatting Kernels}
\label{app: 3D Splatting Kernels}
\subsection{Splatting Condition}

In volume rendering and splatting techniques, a continuous \textbf{3D kernel function} $K(x, y, z)$ is projected onto a 2D plane to produce a continuous \textbf{2D kernel function} $K(x, y)$. This projection is typically performed along one axis (e.g., the $z$-axis) and is represented by the integral:

\begin{equation}
\int_{-\infty}^{\infty} K(x, y, z) \, dz = K(x, y)
\end{equation}

If $K(x, y, z)$ is a \textbf{separable function}, such as:

\begin{itemize}
    \item \textbf{Multiplicative form}: 
    \begin{equation}
    K(x, y, z) = K(x) \cdot K(y) \cdot K(z)
    \end{equation}
    \item \textbf{Additive form}: 
    \begin{equation}
    K(x, y, z) = K(x) + K(y) + K(z)
    \end{equation}
\end{itemize}

It is straightforward to derive the corresponding 2D projection.

However, when we want the kernel to be \textbf{deformable} or to have an \textbf{adjustable shape} (e.g., controlling its spread or sharpness), it may be challenging to compute the integral analytically. For example, consider the radially symmetric kernel:

\begin{equation}
K(x, y, z) = 1 - (x^2 + y^2 + z^2)^\beta
\end{equation}

with $\beta > 0$ and $x^2 + y^2 + z^2 \leq 1$. Computing the integral to obtain $K(x, y)$ can be difficult due to the non-separable nature of the kernel and the dependence on $\beta$.

\subsection{Inverse Splatting}

To address this challenge, we can \textbf{reverse the process} by first defining the desired \textbf{2D kernel} $K(r)$, where $r = \sqrt{x^2 + y^2}$, and then determining whether there exists a corresponding \textbf{3D kernel} $K(R)$, with $R = \sqrt{x^2 + y^2 + z^2}$, such that the projection condition is satisfied:

\begin{equation}
K(r) = \int_{-\infty}^{\infty} K(R) \, dz
\end{equation}

Assuming \textbf{radial symmetry} and \textbf{finite support} (i.e., $r \leq 1$ and $R \leq 1$), we can utilize the \textbf{inverse Abel transform} to derive the 3D kernel $K(R)$ from the given 2D kernel $K(r)$:

\begin{equation}
K(R) = -\frac{1}{\pi} \int_R^1 \frac{dK(r)}{dr} \cdot \frac{1}{\sqrt{r^2 - R^2}} \, dr
\end{equation}

This guarantees that any suitably defined 2D kernel admits a multi-view-consistent 3D counterpart.

\textbf{Example:}

Let’s take the Beta Kernel as an example:

\begin{equation}
K(r) = (1 - r^2)^\beta
\end{equation}

with $\beta > 0$ and $0 \leq r \leq 1$. The radial derivative of $K(r)$ is:

\begin{equation}
\frac{dK(r)}{dr} = -2\beta r (1 - r^2)^{\beta - 1}.
\end{equation}

Substituting into the inverse Abel transform, we obtain the 3D kernel:

\begin{equation}
K(R) = \frac{2\beta}{\pi} \int_R^1 \frac{r (1 - r^2)^{\beta - 1}}{\sqrt{r^2 - R^2}} \, dr.
\end{equation}
This is the multi-view-consistent geometry we observe in 3D space by defining a 2D Beta Kernel in splatted billboards.

\subsection{General Conditions for 2D Kernel Functions}

To ensure that the 2D kernel $K(r)$ corresponds to a valid 3D kernel $K(R)$ via the inverse Abel transform and satisfies the projection condition, the following conditions must be met:

\begin{enumerate}
    \item \textbf{Definition and Range}:
    \begin{itemize}
        \item The kernel is defined for $r \in [0, 1]$.
        \item The kernel values are in the range $K(r) \in [0, 1]$.
    \end{itemize}
    
    \item \textbf{Boundary Conditions}:
    \begin{itemize}
        \item \textbf{Opacity at the center}: 
        \begin{equation}
        \lim_{r \to 0^+} K(r) = 1.
        \end{equation}
        \item \textbf{Transparency at the boundary}: 
        \begin{equation}
        \lim_{r \to 1^-} K(r) = 0.
        \end{equation}
    \end{itemize}
    
    
    \item \textbf{Smoothness and Differentiability}:
    \begin{itemize}
        \item The kernel $K(r)$ is continuously differentiable on $(0, 1)$:
        \begin{equation}
        K(r) \in C^1(0, 1).
        \end{equation}
    \end{itemize}
    
    
    \item \textbf{Integrability}:
    \begin{itemize}
        \item The kernel and its derivative are \textbf{absolutely integrable} over $(0, 1)$:
        \begin{equation}
        \int_0^1 K(r) \, dr < \infty
        \end{equation}
        \begin{equation}
        \int_0^1 \left| \frac{dK(r)}{dr} \right| \, dr < \infty
        \end{equation}
    \end{itemize}
    
    \item \textbf{Existence of Corresponding 3D Kernel}:
    \begin{itemize}
        \item The \textbf{inverse Abel transform} of $K(r)$ yields a valid 3D kernel $K(R)$ that satisfies the projection condition:
        \begin{equation}
        K(R) = -\frac{1}{\pi} \int_R^1 \frac{dK(r)}{dr} \cdot \frac{1}{\sqrt{r^2 - R^2}} \, dr
        \end{equation}
    \end{itemize}
\end{enumerate}

By enforcing the above conditions on any radial 2D function, the inverse Abel transform ensures the existence of a unique radial 3D kernel.
Hence, one may prescribe any 2D splatting kernel and be guaranteed a corresponding, multi-view-consistent 3D geometry—without having to derive the 3D kernel in closed form.  

\begin{table*}[htbp]
    \centering
     \resizebox{0.8\linewidth}{!}{\begin{tabular}{cc|ccc|ccc|ccc|ccc}
    \hline
        
        & \multirow{2}{*}{Scenes}& \multicolumn{3}{c|}{Zip-NeRF} & \multicolumn{3}{c|}{3DGS-MCMC} & \multicolumn{3}{c|}{(Ours,30K,sb=2)} & \multicolumn{3}{c}{(Ours,full,sb=2)}\\
         & &$\text{PSNR}^\uparrow$ &$\text{SSIM}^\uparrow$& $\text{LPIPS}^\downarrow$& $\text{PSNR}^\uparrow$& $\text{SSIM}^\uparrow$& $\text{LPIPS}^\downarrow$& $\text{PSNR}^\uparrow$& $\text{SSIM}^\uparrow$& $\text{LPIPS}^\downarrow$& $\text{PSNR}^\uparrow$& $\text{SSIM}^\uparrow$& $\text{LPIPS}^\downarrow$\\
         \hline
         
         \parbox[t]{2mm}{\multirow{10}{*}{\rotatebox[origin=c]{90}{Mip-NeRF 360  }}}
         & bicycle & 25.80 & 0.769 & 0.208    & 26.15  & 0.81    & 0.18   & 26.04 & 0.804  & 0.169 & 26.10 & 0.805 &  0.170 \\
         & flowers & 22.40 & 0.642 & 0.273    & 22.12  & 0.642   & 0.31  &  22.44 & 0.650  & 0.308 & 22.49 & 0.652 &  0.306  \\
         & garden  & 28.20 & 0.860 & 0.118    & 28.16  & 0.89    & 0.10  & 28.15 & 0.882  & 0.096 & 28.11 & 0.881 &  0.095 \\
         & stump   & 27.55 & 0.800 & 0.193    & 27.80  & 0.82    & 0.19  & 27.56 & 0.815  & 0.184  & 27.50 & 0.817 &  0.178 \\
         & treehill & 23.89 & 0.681 & 0.242   & 23.31  & 0.66    & 0.29   & 23.49 & 0.676  & 0.293 & 23.50 & 0.680 &  0.289 \\
         & room    & 32.65 & 0.925 & 0.196    & 32.48  & 0.94    & 0.25   & 32.83 & 0.941  & 0.168 & 33.03 & 0.942 &  0.165 \\
         & counter & 29.38 & 0.902 & 0.185    & 29.51  & 0.92    & 0.22  & 30.36 & 0.930  & 0.154 & 30.70 & 0.932 &  0.149  \\
         & kitchen & 32.50  & 0.928 & 0.116   & 32.27  & 0.94    & 0.14  & 32.61 & 0.939  & 0.110 & 32.84 & 0.940 &  0.107  \\
         & bonsai  & 34.46 & 0.949 & 0.173    & 32.88  & 0.95    & 0.22   & 33.90 & 0.957  & 0.156 & 34.46 & 0.960 &  0.151  \\
         & Mean    & 28.54 & 0.828 & 0.189    & 28.29  & 0.84    & 0.21   & 28.60 & 0.844  & 0.182 & 28.75 & 0.845 & 0.179 \\
        \hdashline

        \parbox[t]{2mm}{\multirow{9}{*}{\rotatebox[origin=c]{90}{NeRF Synthetic  }}}
        & chair & 34.84 & 0.983 & 0.017 & 36.51 &  0.99 & 0.02 & 36.74 & 0.990 &  0.010 & 36.70 & 0.990 & 0.010  \\
        & drums & 25.84 &  0.944 & 0.050 & 26.29 & 0.95 & 0.04 & 26.78 & 0.958 & 0.033 & 26.81 & 0.958 & 0.032 \\ 
        & ficus & 33.90 &  0.985 & 0.015 & 35.07 & 0.99 & 0.01 & 36.75 & 0.990 &  0.010 & 36.79 & 0.990 & 0.010 \\ 
        & hotdog & 37.14 & 0.984 & 0.020 & 37.82 & 0.99 & 0.02 & 38.85 & 0.988 &  0.015 & 38.95 & 0.988 & 0.016 \\
        & lego & 34.84 & 0.980 & 0.019 & 36.01 & 0.98 & 0.02 & 37.12 & 0.985 &  0.014 & 37.11 & 0.985 & 0.014 \\
        & materials & 31.66 & 0.969 & 0.032 & 30.59 & 0.96 & 0.04 & 31.12 & 0.966 &  0.032 & 31.13 & 0.966 & 0.031 \\ 
        & mic & 35.15 & 0.991 & 0.007 & 37.29 & 0.99 & 0.01  & 37.66 & 0.994 &  0.005 & 37.67 & 0.994 & 0.005  \\ 
        & ship & 31.38 & 0.929 & 0.091 & 30.82 & 0.91 & 0.12 & 32.13 & 0.910 &  0.103 & 32.15 & 0.910 & 0.104  \\
        & Mean & 33.10 & 0.971 & 0.031 & 33.80 & 0.97 & 0.04 & 34.64 & 0.973 &  0.028 & 34.66 & 0.973 & 0.028 \\

        \hdashline
        \parbox[t]{2mm}{\multirow{3}{*}{\rotatebox[origin=c]{90}{TNT}}}
        & truck & - & - & - & 26.11 & 0.89 & 0.14 & 26.44 & 0.897  & 0.110 & 26.59 & 0.901 & 0.106 \\
        & train & - & - & - & 22.47 & 0.83 & 0.24 & 23.13 & 0.839  & 0.185 & 23.11 & 0.845 & 0.174 \\
        & Mean & - & - & - & 24.29 & 0.86 & 0.19 & 24.79 & 0.868  & 0.147 & 24.85 & 0.873 & 0.140 \\
        
        \hdashline
        \parbox[t]{2mm}{\multirow{3}{*}{\rotatebox[origin=c]{90}{DB}}}
        & drjohnson & - & - & - & 29.00 & 0.89 & 0.33 & 29.39 & 0.908  & 0.240 & 29.53 & 0.910 &  0.235  \\
        & playroom & - & - & - & 30.33 & 0.90 & 0.31 & 30.82 & 0.912  & 0.240 &  30.71 & 0.918 &  0.236  \\ 
        & Mean & - & - & - & 29.67 & 0.90 & 0.32 & 30.10 & 0.910  & 0.240 & 30.12 & 0.914 &  0.236 \\
        \hline
    \end{tabular}}
    \caption{Quantitative evaluation of rendering results per scene}
    \label{tab:allmetrics}
\end{table*}

\begin{table*}[htbp]
    \centering
     \resizebox{0.9\linewidth}{!}{\begin{tabular}{cc|cccccccccc}
        & Metrics & bicycle & flowers & garden & stump & treehill & room & counter & kitchen & bonsai & Mean \\
        \hline
        &Number of Primitives& 6.0M & 3.0M & 5.0M & 4.5M & 3.5M & 1.5M & 1.5M & 1.5M & 1.5M & 3.1M \\
        \hdashline
        &Training Time (3DGS-MCMC,30K)& 43m & 26m & 39m & 35m & 29m & 27m & 29m & 27m & 25m & 31m \\
        & Rendering Time (3DGS-MCMC) & 18.77ms & 11.53ms & 15.38ms & 15.75ms & 11.33ms & 8.27ms & 10.37ms & 9.09ms & 8.62ms & 12.12ms \\
        & Rendering FPS (3DGS-MCMC) & 53.30 & 86.72 & 65.00 & 63.54 & 88.26 & 120.86 & 96.43 & 110.06 & 116.01 & 82.46\\
        & Memory (3DGS-MCMC) & 1395.42 MB & 709.54 MB & 1182.56 MB & 1064.30 MB & 827.79 MB & 354.77 MB & 354.77 MB & 354.77 MB & 354.77 MB & 733.19 MB \\
        \hdashline
        &Training Time (30K,sb=2)& 23m & 14m & 20m & 17m & 12m & 14m & 17m & 13m & 12m & 16m\\
        &Training Time (full,sb=2)& 24m & 11m & 25m & 14m & 12m & 19m & 32m & 28m & 32m & 22m\\
        &Training Steps (full,sb=2)& 31.0K & 26.5K & 25.0K & 25.5K & 25.5K & 48.5K & 64.0K & 61.5K & 84.5 & 43.5K\\
        &Rendering Time (sb=2)& 12.70ms & 8.13ms & 8.71ms & 11.33ms & 7.42ms & 5.03ms & 7.69ms & 7.04ms & 5.04ms & 8.12ms\\
        &Rendering FPS (sb=2)& 78.74 & 122.93 & 114.71 & 88.21 & 134.82 & 198.97 & 130.01 & 141.93 & 198.34 & 123.09\\
        &Memory (sb=2) & 686.65 MB & 343.32 MB & 572.21 MB & 514.98 MB & 400.54 MB & 171.66 MB & 171.66 MB & 171.66 MB & 171.66 MB & 356.04 MB \\
        \hline
    \end{tabular}}
    \caption{Quantitative evaluation of training time and rendering efficiency per scene}
    \label{tab:fulltraining}
\end{table*}

\section{Per-scene Results}
\label{app:additional}

Our per-scene evaluation on the Mip-NeRF 360 dataset shows that, on diffuse scenes (\eg, “bicycle,” “stump,” and “flowers”), Deformable Beta Splatting achieves the comparable performance of prior works, while on scenes with strong specular reflections (\eg, “counter”), it surpasses 3DGS-MCMC and Zip-NeRF by over 1 dB in PSNR. This gain demonstrates the ability of DBS to faithfully reconstruct intricate specular lighting in challenging scenarios.

\section{Kernel Comparison}

\begin{figure}[htbp]
    \centering
    \includegraphics[width=0.8\linewidth]{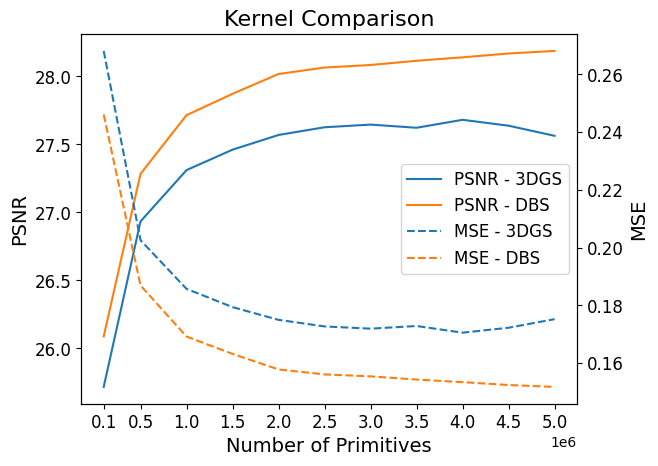}
    \caption{Reconstruction quality on the Mip-NeRF 360 without view-dependent color-encodings. Deformable Beta Splatting (DBS) consistently outperforms 3D Gaussian Splatting (3DGS), confirming the stronger representation ability of the Beta kernel.}
    \label{fig:kernel-comparison}
\end{figure}

In addition to \Cref{tab:ablation}, we isolate the effect of the splatting kernel through an ablation study, disabling all view-dependent color terms (\texttt{sh=0}, \texttt{sb=0}) and comparing the standard Gaussian kernel (3DGS) to our deformable Beta kernel (DBS). Both models are trained on the Mip-NeRF 360 training set and evaluated on test views under MCMC densification schedules.

\Cref{fig:kernel-comparison} plots PSNR (left axis) and MSE (right axis) as a function of the total number of primitives. Although both kernels improve with more primitives, the Gaussian baseline plateaus around 1.5 M, while the Beta kernel continues to yield gains. Across the entire budget (0.1 M–5 M primitives), DBS consistently achieves higher PSNR and lower MSE, demonstrating that the deformable Beta kernel alone provides superior geometric modeling capacity.

\section{Kernel-Agnostic Distribution-Preserved Densification}
\label{app:densification}
\subsection{Detailed Proof}

\begin{figure*}[htbp]
    \centering
    \includegraphics[width=0.83\linewidth]{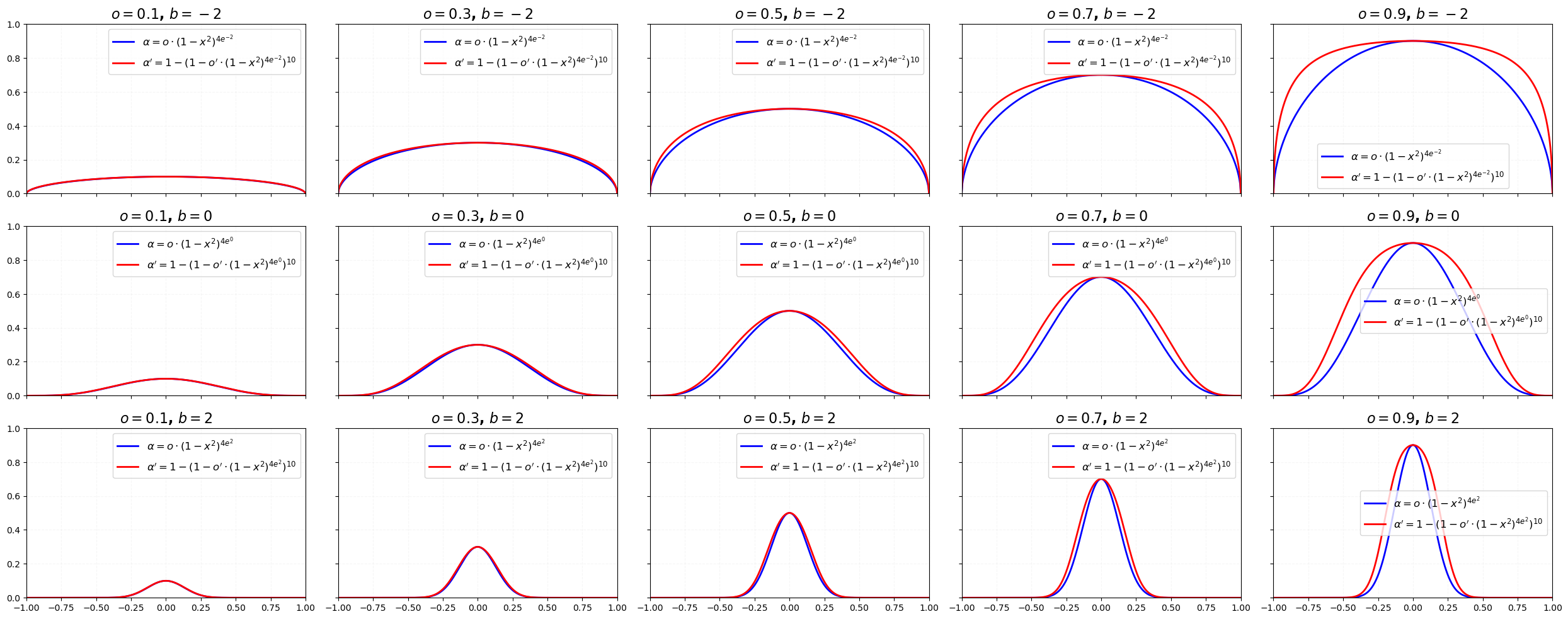}
    \caption{Kernel-Agnostic Distribution-Preserved Densification illustration.}
    \label{fig:densification}
\end{figure*}

Assume a primitive defined by a splatting kernel $f(x)$ with opacity $o$, and it will be cloned $N$ times, Its contribution at a pixel x is given by:
\begin{equation}
    \alpha = o\cdot f(x)
\end{equation}
When cloning this primitive into $N$ copies, each with new opacity $o'$, the combined contribution becomes:
\begin{equation}
    \alpha' = 1 - (1-o'\cdot f(x))^N
\end{equation}
To preserve its distribution, we require $\alpha=\alpha'$, such that
\begin{equation}
    o\cdot f(x) = 1 - (1-o'\cdot f(x))^N
    \label{eq:densify}
\end{equation}
Ignoring the kernel momentarily, the condition simplifies to:
\begin{equation}
    o' = 1 - \sqrt[N]{1 - o},
\end{equation}
which aligns with established proof~\cite{bulò2024revisingdensificationgaussiansplatting, kheradmand20243dgaussiansplattingmarkov}.
Incorporating $f(x)$ complicates preservation beyond the primitive center, especially at boundaries. Prior approaches like 3DGS-MCMC handled boundary preservation via scale adjustments tailored to Gaussian kernels, which are complex and not directly applicable to deformable kernels.

We will show that given a small opacity value, only the opacity adjustment is effective and valid regardless the densified number and kernel function.

For small $o$, we approximate $o'$ using a Taylor expansion $\sqrt[N]{1 - o} \approx 1 - \frac{o}{N}
$, such that 
\begin{equation}
    o' = 1 - \sqrt[N]{1 - o} \approx \frac{o}{N}
\end{equation}

Now consider the densified distribution:
\begin{equation}
    \alpha' = 1 - \Big(1 - o' \cdot f(x)\Big)^N\approx 1 - \Big(1 - \frac{o}{N} f(x)\Big)^N.
\end{equation}

Since $\frac{o}{N} f(x)$ is also small, applying the binomial approximation:
\begin{equation}
\Big(1 - \frac{o}{N} f(x)\Big)^N \approx 1 - N \cdot \frac{o}{N} f(x) + \mathcal{O}(o^2) = 1 - o \, f(x) + \mathcal{O}(o^2),
\end{equation}
 we find:
\begin{equation}
\alpha' \approx 1 - \big[1 - o \, f(x) + \mathcal{O}(o^2)\big] = o \, f(x) + \mathcal{O}(o^2).
\end{equation}

This matches the original distribution:
\begin{equation}
\alpha = o \cdot f(x),
\end{equation}
up to first-order in $o$.

The error term $\mathcal{O}(o^2)$ becomes negligible as $o$ decreases. Therefore, the difference between original distribution and densified distribution diminishes for smaller $o$ regardless of copy times $N$ and splatting kernel $f(x)$.

\subsection{Validity of the Small Opacity Assumption}

A natural question arises regarding the role of the small opacity prior used in our framework. At first glance, encouraging primitives to remain semi-transparent may appear counterintuitive: in surface-based representations, higher opacity is typically desirable to achieve clean geometry.
However, our approach lies in Volumetric Density field. Volumetric methods—such as NeRF and 3DGS—fundamentally differ from surface‐based techniques. In a volumetric framework, each primitive contributes only a small opacity, and the final image emerges from the accumulation of many low‐opacity contributions. Within this paradigm, enforcing high per-primitive opacity can improve geometric fidelity and facilitate mesh extraction, but contradicts the core assumptions of smooth transmittance accumulation and degrades visual quality. Notable examples of this trade-off include NeuS~\cite{wang2021neus} vs NeRF~\cite{mildenhall2021nerf} and SuGaR~\cite{guédon2023sugarsurfacealignedgaussiansplatting} vs 3DGS~\cite{3dgs}. In contrast, applying a moderate opacity prior aligns with the volumetric principle, supports distribution-preserving densification, and further stabilizes the optimization process.

To quantify the impact of the opacity regularization, we analyze its effect both theoretically and empirically. In \Cref{fig:densification}, we visualize the original distribution (blue) and its densified distribution (red) as defined in \Cref{eq:densify}, across different kernel shapes. As opacity decreases, the discrepancy between the two distributions diminishes, with near-perfect alignment observed when the opacity approaches 0.1. This behavior supports our theoretical analysis: under the small-opacity regime, the densified distribution accurately preserves the original, regardless of the number of clones or the kernel function used.

\begin{table}[htbp]
    \centering
    \begin{tabular}{ccc}
        \hline
        $\lambda_o$    & PSNR & Opacity \\ 
        \hline
        0.000  & 27.81     & 0.60 $\pm$ 0.05 \\
        0.001  & 28.31     & 0.26 $\pm$ 0.05 \\
        0.005  & 28.61     & 0.14 $\pm$ 0.04 \\
        \textbf{0.010}  & \textbf{28.70}     & \textbf{0.10 $\pm$ 0.03} \\
        0.050  & 28.69     & 0.04 $\pm$ 0.02 \\
        0.100  & 25.07     & 0.02 $\pm$ 0.02 \\
        \hline
    \end{tabular}
    \caption{Mean PSNR and Opacity Distribution on the Mip-360 dataset for different $\lambda_o$ values.}
    \label{tab:psnr_opacity}
\end{table}

Complementing this, we conduct an ablation study on the opacity regularization weight $\lambda_o$ across the Mip-NeRF 360 dataset. We report mean PSNR on held-out test views along with the mean and standard deviation of primitive opacities. As shown in \Cref{tab:psnr_opacity}, omitting the opacity term causes the densification step to naively clone primitives, resulting in over-saturated reconstructions and reducing visual quality. Introducing the opacity prior yields consistent gains in PSNR by reducing both the average opacity and its variance. However, excessively large $\lambda_o$ values overly suppress opacity and drive primitives toward excessive transparency, again causing damage to the quality of reconstruction. Empirically, we observe that setting $\lambda_o=0.01$ produces an average opacity around 0.1 and achieves peak PSNR, thereby validating our small opacity assumption.

\begin{table*}[htbp]
    \centering
     \resizebox{0.8\linewidth}{!}{
     \begin{tabular}{c|ccccccc}
    \hline
         Method 
          & $\text{PSNR}^\uparrow$ 
          & $\text{SSIM}^\uparrow$ 
          & $\text{LPIPS}^\downarrow$ 
          & $\text{Num (M)}^\downarrow$
          & $\text{Size (MB)}^\downarrow$
          & Compression Time (s)$^\downarrow$ 
          & Compression Ratio$^\uparrow$\\
         \hline
         3DGS 
           & 27.20 
           & 0.815 
           & 0.214 
           & 3.35
           & 752.74 
           & – 
           & 1 \\
         DBS‑full 
           & \cellcolor{red!25}28.75 
           & \cellcolor{red!25}0.845 
           & \cellcolor{red!25}0.179 
           & \cellcolor{yellow!25}3.11
           & 356.04 
           & – 
           & 2.11 \\
         DBS‑full (compressed) 
           & \cellcolor{orange!25}28.60 
           & \cellcolor{orange!25}0.840 
           & \cellcolor{orange!25}0.182 
           & \cellcolor{yellow!25}3.11
           & 63.48 
           & \cellcolor{yellow!25}41.70 
           & 11.86 \\
         DBS‑1M 
           & \cellcolor{yellow!25}28.42 
           & \cellcolor{yellow!25}0.831 
           & \cellcolor{yellow!25}0.205 
           & \cellcolor{orange!25}1.00
           & 114.44 
           & – 
           & 6.58 \\
         DBS‑1M (compressed) 
           & 28.32 
           & 0.828 
           & 0.207 
           & \cellcolor{orange!25}1.00
           & \cellcolor{orange!25}22.32 
           & \cellcolor{orange!25}14.28 
           & \cellcolor{orange!25}33.72 \\
         DBS‑490K 
           & 28.03 
           & 0.816 
           & 0.232 
           & \cellcolor{red!25}0.49
           & \cellcolor{yellow!25}56.08 
           & – 
           & \cellcolor{yellow!25}13.42 \\
         DBS‑490K (compressed) 
           & 27.80 
           & 0.806 
           & 0.237 
           & \cellcolor{red!25}0.49
           & \cellcolor{red!25}11.37 
           & \cellcolor{red!25}9.15 
           & \cellcolor{red!25}66.20 \\
        \hline
     \end{tabular}
     }
    \caption{Quantitative evaluation of visual quality and model size on the Mip-360 dataset. More compression comparisons are available at \href{http://3DGS.zip}{3DGS.zip}~\cite{bagdasarian20253dgszipsurvey3dgaussian}.}
    \label{tab:compression}
\end{table*}

\section{Compression}
\label{app:compression}
In \Cref{tab:metrics_nonerfsynthetic} and \Cref{tab:efficiency}, we configured DBS-full to use the similar number of primitives as vanilla 3DGS (3.35 M) for a fair comparison. Thanks to the MCMC framework, we can precisely control the target primitive count—so we also evaluate two "pruned" variants, DBS-1M (1 M primitives) and DBS-490 (490 K primitives), to explore the trade-off between model size and visual quality. 
After training, we apply a post-training compression pipeline adapted from~\cite{morgenstern2024compact} to further reduce each model’s on-disk footprint.
Specifically, we firstly sort and reorder the parameters so that similar values are adjacent, maximizing the efficiency of lossless PNG compression. Next, we quantize most parameters from 32-bit floats down to 8-bit, allowing us to save them into PNG images. However, because we observed that reducing the precision of the 3D position to 8 bits causes a severe degradation in rendering quality, we preserve those values at full 32-bit precision, distributing them across four separate images, while all other parameters are stored as 8-bit PNGs.

We conduct these experiments on the Mip-360 dataset and report mean metrics across nine scenes in Table \ref{tab:compression}.
As the model is pruned from 3.11 M primitives (DBS-full) to 1 M (DBS-1M) and then to 0.49 M (DBS-490K), we observe a graceful degradation in rendering fidelity: mean PSNR decreases from 28.75 dB to 28.42 dB to 28.03 dB, SSIM from 0.845 to 0.831 to 0.816, and LPIPS increases from 0.179 to 0.205 to 0.232. In parallel, the raw on-disk footprint contracts substantially—from 356 MB (2.11x relative to vanilla 3DGS) to 114 MB (6.58x) to 56 MB (13.42x). Applying uniform post-training PNG compression further amplifies these storage savings by approximately 5x–6x: final sizes become 63 MB, 22 MB, and 11 MB, corresponding to overall compression ratios of 11.9x, 33.7x, and 66.2x relative to 3DGS. Crucially, this additional compression incurs only minor quality losses (PSNR drops of 0.15 dB, 0.10 dB, and 0.23 dB) and remains computationally practical, requiring only 41.7 s, 14.3 s, and 9.2 s per scene, respectively. The naively compressed DBS-490K matches vanilla 3DGS’s visual quality in only $\sim$10 MB.

\section{Hyperparameters}
\label{app:hyperparameters}

The training process is governed by several hyperparameters. The initial learning rate for position optimization is set to $\text{lr}_{\mathbf{\mu}}^{\text{init}}=0.00016$, which decays to a final value of $\text{lr}_{\mathbf{\mu}}^{\text{final}} = 0.0000016$. A delay multiplier of $0.01$ is applied, with a maximum number of steps for position learning rate decay of $30{,}000$. The base color learning rate is $\text{lr}_{c_0} = 0.0025$, and the same value is used for the Spherical Beta parameters learning rate ($\text{lr}_{b_m,\hat{R}_m,c_m} = 0.0025$). The opacity learning rate is relatively high, set to $\text{lr}_{o} = 0.05$, while the beta learning rate is $\text{lr}_{b} = 0.001$. The scaling and rotation learning rates are $\text{lr}_{\mathbf{s}} = 0.005$ and $\text{lr}_{\mathbf{q}} = 0.001$, respectively.

Furthermore, the loss function is regularized using a DSSIM loss term with a weight of $\lambda_{\text{SSIM}} = 0.2$. Densification occurs every $100$ iterations, starting from $500$ and continuing until $25{,}000$. The background is not randomized during training, and a learning rate noise of $\lambda_{\text{lr}} = 5 \times 10^4$ is used. The regularization terms for scale and opacity are $\lambda_{\Sigma} = 0.01$ and $\lambda_o = 0.01$, respectively.
\end{document}